\documentclass[pdflatex,sn-mathphys-num]{sn-jnl}

\usepackage{graphicx}%
\usepackage{multirow}%
\usepackage{amsmath,amssymb,amsfonts}%
\usepackage{amsthm}%
\usepackage{mathrsfs}%
\usepackage[title]{appendix}%
\usepackage{xcolor}%
\usepackage{textcomp}%
\usepackage{manyfoot}%
\usepackage{booktabs}%
\usepackage{algorithm}%
\usepackage{algorithmicx}%
\usepackage{algpseudocode}%
\usepackage{listings}%
\usepackage{wrapfig} 
\usepackage{subcaption}
\usepackage{multirow}
\usepackage{makecell}
\usepackage{longtable}

\theoremstyle{thmstyleone}%
\theoremstyle{thmstyletwo}%

\theoremstyle{thmstylethree}%

\begin{document}

\title[Article Title]{Skill-Nav: Enhanced Navigation with Versatile Quadrupedal Locomotion via Waypoint Interface}

\author[1,3]{\fnm{Dewei} 
\sur{Wang}}\email{dweik0607@gmail.com}

\author*[3]{\fnm{Chenjia} 
\sur{Bai}}\email{baicj@chinatelecom.cn}

\author[2]{\fnm{Chenhui} 
\sur{Li}}\email{lichenhui@pjlab.org.cn}

\author[3]{\fnm{Jiyuan} 
\sur{Shi}}\email{shijy15@chinatelecom.cn}


\author[2]{\fnm{Yan} 
\sur{Ding}}\email{yding25@binghamton.edu}

\author[3]{\fnm{Chi} 
\sur{Zhang}}\email{zhangc120@chinatelecom.cn}

\author*[2,4]{\fnm{Bin} 
\sur{Zhao}}\email{zhaobin@pjlab.org.cn}

\affil[1]{\orgdiv{School of Information Science and Technology}, \orgname{University of Science and Technology of China}, \orgaddress{\city{Hefei}, \country{China}}}

\affil[2]{\orgname{Shanghai Artificial Intelligence Laboratory}, \orgaddress{\city{Shanghai}, \country{China}}}

\affil[3]{\orgdiv{Institute of Artificial Intelligence (TeleAI)}, \orgname{China Telecom Corp Ltd}, \orgaddress{\city{Shanghai}, \country{China}}}


\affil[4]{\orgdiv{School of Artificial Intelligence, Optics and Electronics (iOPEN)}, \orgname{Northwestern Polytechnical University}, \orgaddress{\city{Xi’an}, \country{China}}}


\abstract{Quadrupedal robots have demonstrated exceptional locomotion capabilities through Reinforcement Learning (RL), including extreme parkour maneuvers. However, integrating locomotion skills with navigation in quadrupedal robots has not been fully investigated, which holds promise for enhancing long-distance movement capabilities. In this paper, we propose Skill-Nav, a method that incorporates quadrupedal locomotion skills into a hierarchical navigation framework using waypoints as an interface. Specifically, we train a waypoint-guided locomotion policy using deep RL, enabling the robot to autonomously adjust its locomotion skills to reach targeted positions while avoiding obstacles. Compared with direct velocity commands, waypoints offer a simpler yet more flexible interface for high-level planning and low-level control. Utilizing waypoints as the interface allows for the application of various general planning tools, such as large language models (LLMs) and path planning algorithms, to guide our locomotion policy in traversing terrains with diverse obstacles. Extensive experiments conducted in both simulated and real-world scenarios demonstrate that Skill-Nav can effectively traverse complex terrains and complete challenging navigation tasks.}


\maketitle

\section{Introduction}\label{sec1}

\begin{wrapfigure}{r}{0.48\textwidth}
    \vspace{-5mm}
    \centering
    \includegraphics[width=0.48\textwidth]{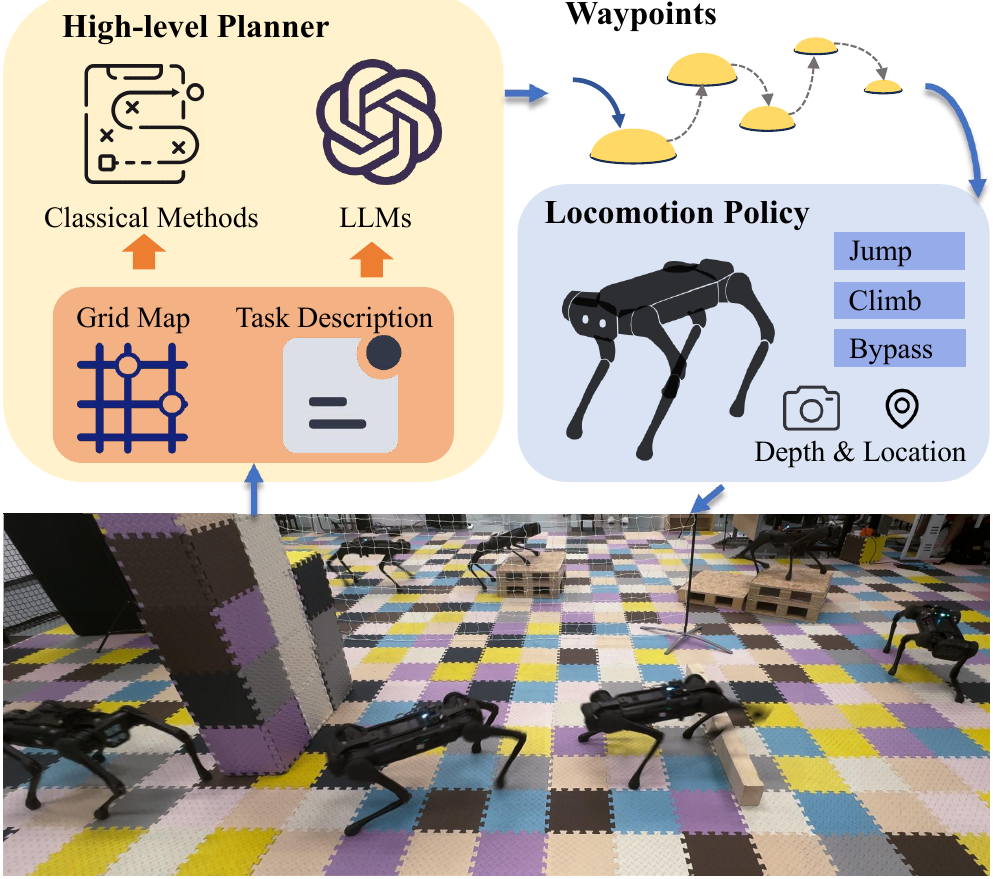}
    \caption{Overview of Skill-Nav: the high-level planner guides the low-level locomotion policy using waypoints}
    \label{fig:method_overview}
    \vspace{-5mm}
\end{wrapfigure}

Artificial Intelligence (AI) algorithms, represented by deep learning, have made outstanding advances in various domains like computer vision \cite{jiang2018deep, gao2015learning, tao2006human, han2015two, tao2008bayesian}. Robots with AI also increasingly fulfill crucial roles in various tasks, such as planetary exploration and complex manipulation \cite{arm2023scientific, 2022cerberus, qu2025spatialvla, song2025hume,an2025aiflowperspectivesscenarios, wang2025more}. 
Quadrupedal robots offer unparalleled flexibility and the capability to navigate challenging terrains. Fundamental to their interaction with the real world are advanced locomotion and navigation systems. In terms of locomotion, Model Predictive Control (MPC)-based methods \cite{2018MPC2} model environmental dynamics to achieve precise control. Reinforcement Learning (RL)-based methods \cite{2020challengingterrain, 2021rma, 2022inthewild, 2022leggedgym}, trained in simulated environments like Isaac Gym \cite{isaacgym}, utilize curriculum learning \cite{bengio2009curriculum} and domain randomization to enhance robustness and versatility. However, most locomotion policies that track velocity commands face significant tracking errors, which undermine their suitability for navigation tasks. Existing navigation approaches often use a hierarchical architecture where a high-level planner generates guidance signals based on environmental information like maps \cite{2024viplanner, vinl, efficientsafenav, barkour, xu2024optimal}. While some studies \cite{ topnav, coupling, karnan2023sterling, 2021learningembeddings} have incorporated robot feedback to aid navigation, the integration of a versatile locomotion policy with a general planner remains largely unexplored.

Recent learning-based policies have equipped quadrupedal robots with the capabilities to execute extreme locomotion and obstacle avoidance \cite{extreme, robotparkour, 2024High-Obstacle, agilebufsafe}. However, integrating these capabilities into navigation systems remains a huge challenge. As identified in \cite{wheeled-legged}, conventional navigation planners that rely on velocity commands often suffer from tracking errors in the low-level controller. Fully learning-based approaches \cite{barkour, Anymalparkour} have developed locomotion-specific high-level policies but encounter limitations in generalization. We highlight that using waypoints as an interface offers substantial benefits over velocity commands for bridging the low-level controller with the high-level planner. Firstly, a waypoint-guided locomotion policy facilitates diverse autonomous gaits and reduces dependence on detailed planning. Secondly, waypoints provide sparse, easily generated signals that are more adaptable for a variety of high-level planners. With precise location data, the high-level planner can deliver accurate waypoint instructions, while the low-level controller, guided by waypoints, is less susceptible to the tracking errors that trouble velocity-guided controller.

In this work, we introduce Skill-Nav, designed to navigate complex environments leveraging quadrupedal locomotion skills. Figure \ref{fig:method_overview} depicts the Skill-Nav framework, which consists of a low-level policy capable of traversing diverse terrains guided by waypoints, and a high-level planner that generates these waypoints based on coarse-grained environmental information and the capabilities of the policy. Our approach trains the low-level locomotion controller via RL, using 2D positions relative to the base frame as waypoint commands. We employ two training scenarios: a fixed waypoints scenario
for pre-training, and a random waypoints scenario 
to introduce increased complexity. For the high-level planning, we utilize either a classical path planning algorithm or a large language model (LLM) to enhance system generalization. Given the powerful locomotion capabilities of the low-level controller, our high-level planner does not need to account for intricate terrain details while planning.

The primary contributions of this work are as follows:

\begin{itemize} 
\item We design two training scenarios for a waypoint-guided quadrupedal locomotion policy that enables the robot to perform versatile locomotion. \item We introduce Skill-Nav, which utilizes waypoints as an interface to integrate our locomotion policy with a general high-level planner based on coarse-grained environmental information. 
\item Experiments confirm that Skill-Nav effectively manages navigation tasks in challenging scenarios by leveraging diverse quadrupedal locomotion skills. 
\end{itemize}

\section{Related Work}\label{sec2}

\subsection{Quadrupedal Locomotion}

Classical quadrupedal locomotion methods typically rely on MPC, which demands high-precision dynamic modeling \cite{2018MPC2, 2023MPC1}. Alternatively, RL-based methods have shown remarkable performance in traversing complex terrains using simulated learning and proprioceptive information \cite{2022inthewild, 2022leggedgym, 2023walkthese, 2024rapid, shi2024robust, 2024HIM, 2024limitedperception, shi2024rethinking}. The integration of external sensors such as cameras and LiDAR with RL-based methods has further enhanced performance \cite{2022lquaTransformers, miki2024confined, 2023egocentric, walkbysteering}. Additionally, hybrid approaches that combine the strengths of MPC and RL have improved both accuracy and generalization of locomotion \cite{DTC, kang2023rl+}. Advanced maneuvers, including robot parkour, are now possible through RL and carefully designed reward functions \cite{extreme, robotparkour}.

Recent studies have shown that using target poses as commands is more effective than velocity commands in enhancing autonomous robotic behaviors \cite{advancedskills, resilient, 2023risky, 2024learningdiverse, faust2018prmrl}. To integrate locomotion with local navigation, an end-to-end quadrupedal locomotion policy has been developed, enabling autonomous selection of behaviors to reach local targets \cite{advancedskills}. Challenges such as sparse footholds and risky terrains are addressed through a two-stage training and exploration strategy that incorporates navigation-based formulations \cite{2023risky}. An asymmetric actor-critic architecture paired with a recurrent neural network (RNN) is employed to manage sensor failures \cite{resilient}. Additionally, waypoint-based policies have been developed to generate diverse behaviors through multi-constraint optimization \cite{2024learningdiverse}. Our approach follows the formulation of local navigation to train the robot in two simulated scenarios to enhance its locomotion capabilities.

\subsection{Hierarchical Navigation Architecture}

Navigation systems typically employ a hierarchical architecture, divided into a high-level planner providing guidance signals and a low-level controller that follows these signals \cite{xiao2022motion}. Classical navigation methods often rely on complex model-based or handcrafted algorithms, which frequently lack generalization capabilities in real-world applications \cite{wellhausen2021rough}. An alternative, employing a learning-based method as the low-level controller, offers higher compatibility compared to end-to-end navigation approaches \cite{pfeiffer2017perception, zhu2017target, shah2023vint}. For example, IntentionNet \cite{IntentionNet} uses classical planning algorithms for high-level planning and trains a low-level controller through imitation learning, with intentions serving as the interface. Other research has focused on integrating visual and proprioceptive information into path planners. VP-Nav designed a safety advisor to ensure safe command execution \cite{coupling}, while TOP-Nav developed a terrain estimator that combines a visual estimator with online corrections \cite{topnav}. However, these methods primarily navigate terrains without utilizing advanced locomotion skills. Fully learned hierarchical approaches like Barkour and ANYmal Parkour demonstrate impressive quadrupedal navigation using diverse locomotion skills and elevation maps but often face challenges with generalization and the complexity of training due to tightly coupled modules \cite{barkour, Anymalparkour}.

Our goal is to integrate our versatile locomotion policy with general planners to navigate complex terrains effectively, without a complicated training process. Recent studies suggest that both classical planning algorithms and LLMs can serve as effective planners, inspiring us to use them to generate waypoints and guide our locomotion policy \cite{2024quadrupedgpt, commonsense, IntentionNet}.

\begin{figure*}[t]
    \centering
    \includegraphics[width=\textwidth]{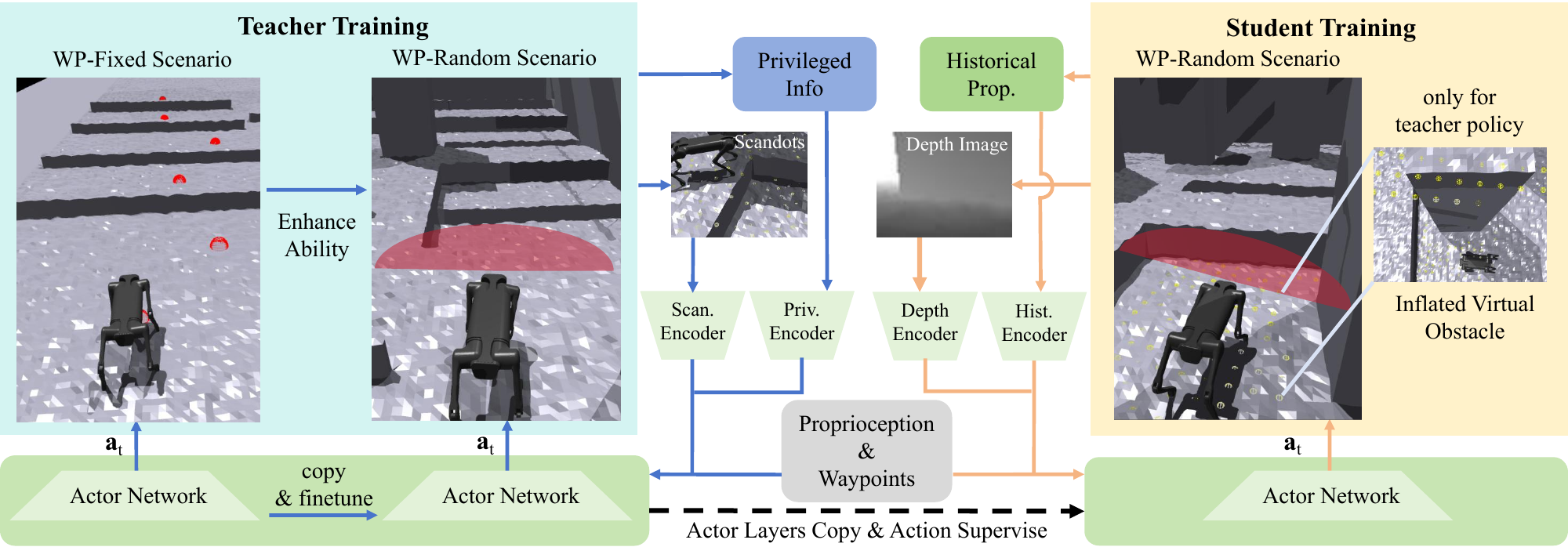}
    \caption{Training pipeline for our locomotion policy. The teacher policy is sequentially trained in both \emph{WP-Fixed} and \emph{WP-Random} scenarios using privileged information (\textit{left}), and its behaviors are distilled into a deployable student policy (\textit{right}).}
    \vspace{-0.4cm}
    \label{fig:training overview}
\end{figure*}

\section{Method}\label{sec4}

The locomotion policy enables a quadruped robot to sequentially traverse through a series of waypoints while performing complex maneuvers such as jumping, climbing, and avoiding obstacles. This policy is optimized using Proximal Policy Optimization (PPO) \cite{2017ppo}. We basically follow the paradigm from previous works \cite{2022leggedgym, extreme}, which utilize a teacher-student architecture to achieve superior locomotion performance. For quadrupedal navigation, we employ off-the-shelf general planners that are tasked with generating waypoints for the locomotion policy.
The high-level planner utilizes coarse-grained environmental information to devise routes, demonstrating robust generalization capabilities.

\subsection{Problem Definition}

The RL-based locomotion control is formulated as a Markov Decision Process (MDP), defined by the tuple \((\mathcal{S}, \mathcal{A}, \mathcal{P}, R)\). Here, \(\mathcal{S}\) represents the state space, \(\mathcal{A}\) the action space, \(\mathcal{P}(\cdot|s, a)\) the state transition function, and \(R: \mathcal{S} \times \mathcal{A} \rightarrow \mathbb{R}\) the reward function. For the teacher policy, the state space \(\mathbf{s}_t\) encompasses several components: proprioception \(\mathbf{p}_t\), estimated robot base linear velocity \(\mathbf{\hat{v}}_t\), \emph{waypoint command} \(\mathbf{w}_t\), encoded privilege information \(\mathbf{e}_t\) and terrain scandots \(\mathbf{m}_t\). The state \(\mathbf{s}_t\) is defined by the following equation:

\vspace{-0.15cm}
\begin{equation}
\mathbf{s}_t = [\mathbf p_t, \mathbf{\hat{v}}_t, \mathbf w_i, \mathbf e_t, \mathbf m_t],
\end{equation}
the proprioceptive information, denoted as \(\mathbf{p}_t\), includes the robot's base angular velocity, Euler angles derived from the inertial measurement unit (IMU), the gravity vector in the robot's frame, joint positions and velocities, the last action taken, and feet contact statuses. For the student policy, historical proprioception and depth encoders are utilized to reconstruct \(\mathbf{e}_t\) and \(\mathbf{m}_t\) respectively, as outlined in \cite{extreme}. The action space of the policy is defined as incremental adjustments to the default joint positions.

The high-level planner is responsible for generating a series of waypoints that guide the quadrupedal robot toward a distant goal. The planning process employed by the high-level planner is encapsulated by the following equation:
\vspace{-0.15cm}
\begin{equation}
\mathcal{W} = \mathcal{H}(\mathbf{M} ,p_e, p_s),
\vspace{-0.2cm}
\end{equation}
in which \( p_e \) and \( p_s \) denote the goal and starting position, respectively. The set \( \mathcal{W} = (\mathbf{w_1, w_2, \ldots, w_i, \ldots, w_n}) \) comprises a sequence of waypoints connecting the start to the goal. \( \mathbf{M} \) represents the coarse-grained environmental information, which varies depending on \( \mathcal{H} \). Here, \( \mathcal{H} \) may be either a classical path planning algorithm or a LLM planner.

\subsection{Waypoint-guided Locomotion Policy}

Fig.~\ref{fig:training overview} illustrates the entire training procedure. The teacher policy is initially trained in a fixed waypoints (\emph{WP-Fixed}) scenario and subsequently fine-tuned in a random waypoints (\emph{WP-Random}) scenario. In contrast, the student policy is exclusively trained in the \emph{WP-Random} scenario. Our policy outperforms those trained solely in one of the aforementioned scenarios, as demonstrated in Sec. \ref{loco-comp}.

\subsubsection{WP-Fixed scenario for skill learning} 

In the \emph{WP-Fixed} scenario, the policy network is randomly initialized and trained to enable the quadrupedal robot to perform basic locomotion skills such as climbing onto boxes and avoiding obstacles while moving forward. 
\begin{wrapfigure}{r}{0.6\textwidth}
    \centering
    \begin{minipage}{0.15\textwidth}
        \centering
        \includegraphics[scale=0.055]{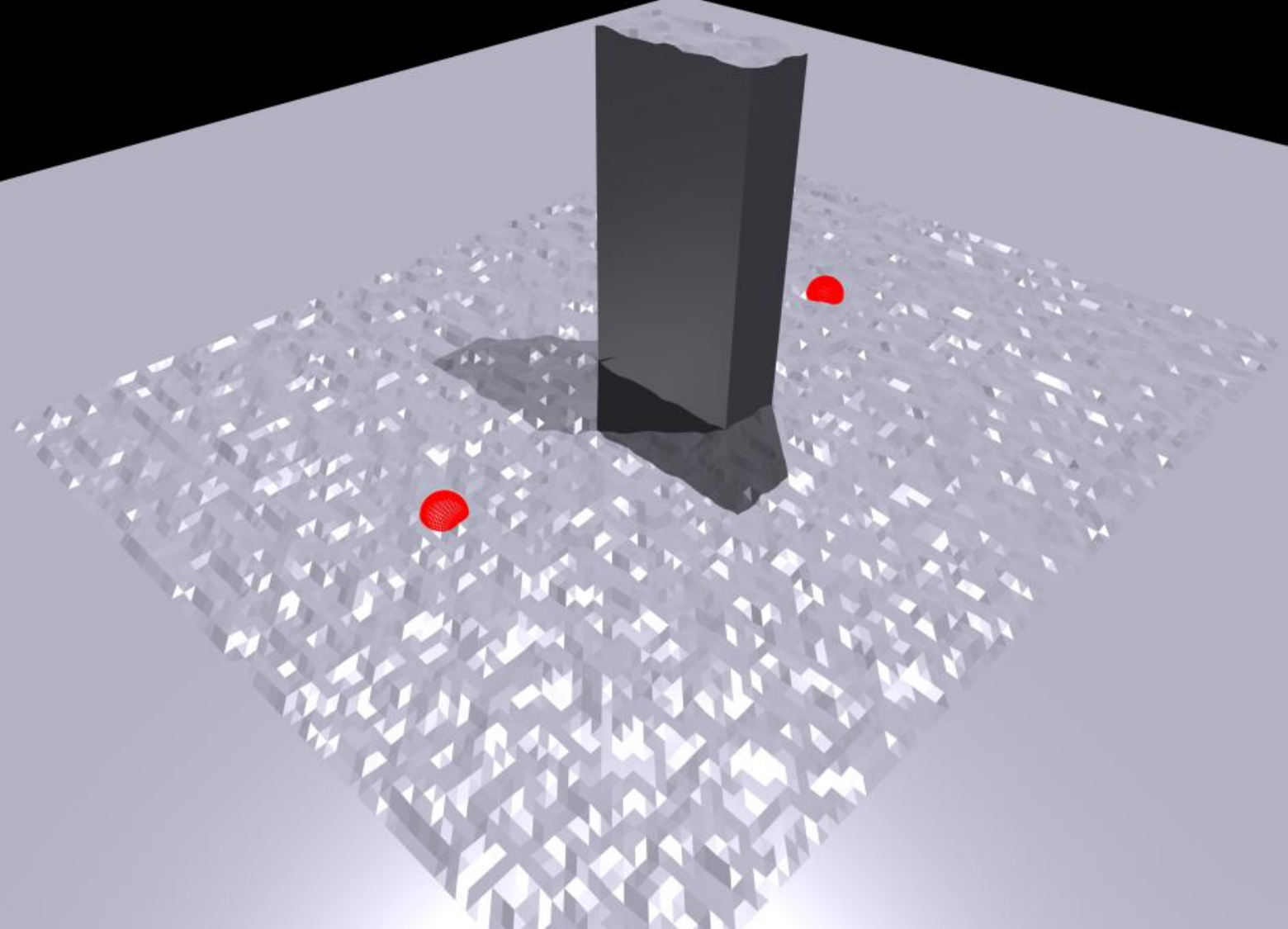}
        \subcaption{Obstacle}
        \label{obstacle}
    \end{minipage}\hspace{-2mm}
    \begin{minipage}{0.15\textwidth}
        \centering
        \includegraphics[scale=0.055]{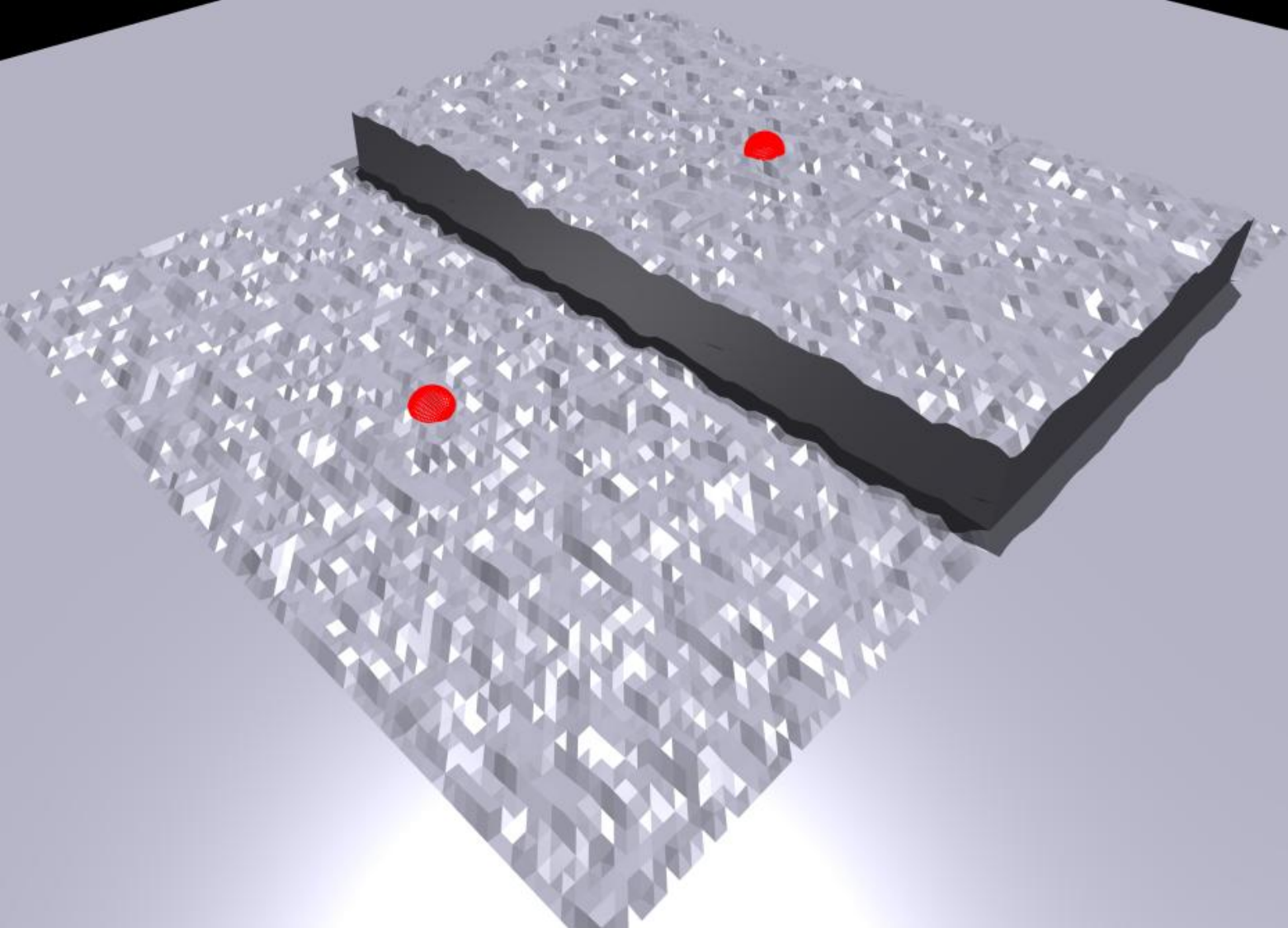}
        \subcaption{Box}
        \label{box}
    \end{minipage}\hspace{-2mm}
    \begin{minipage}{0.15\textwidth}
        \centering
        \includegraphics[scale=0.055]{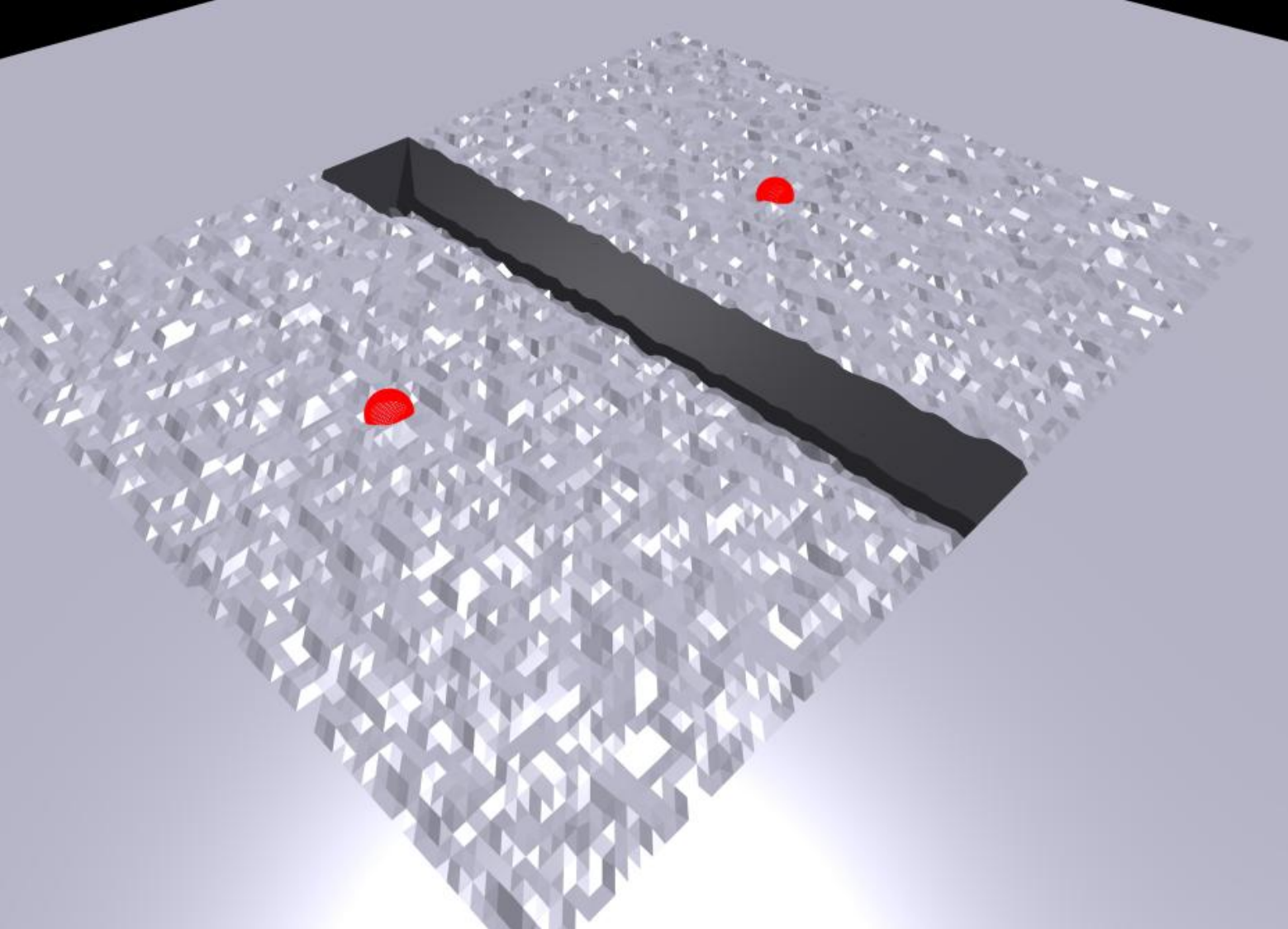}
        \subcaption{Gap}
        \label{gap}
    \end{minipage}\hspace{-2mm}
    \begin{minipage}{0.15\textwidth}
        \centering
        \includegraphics[scale=0.055]{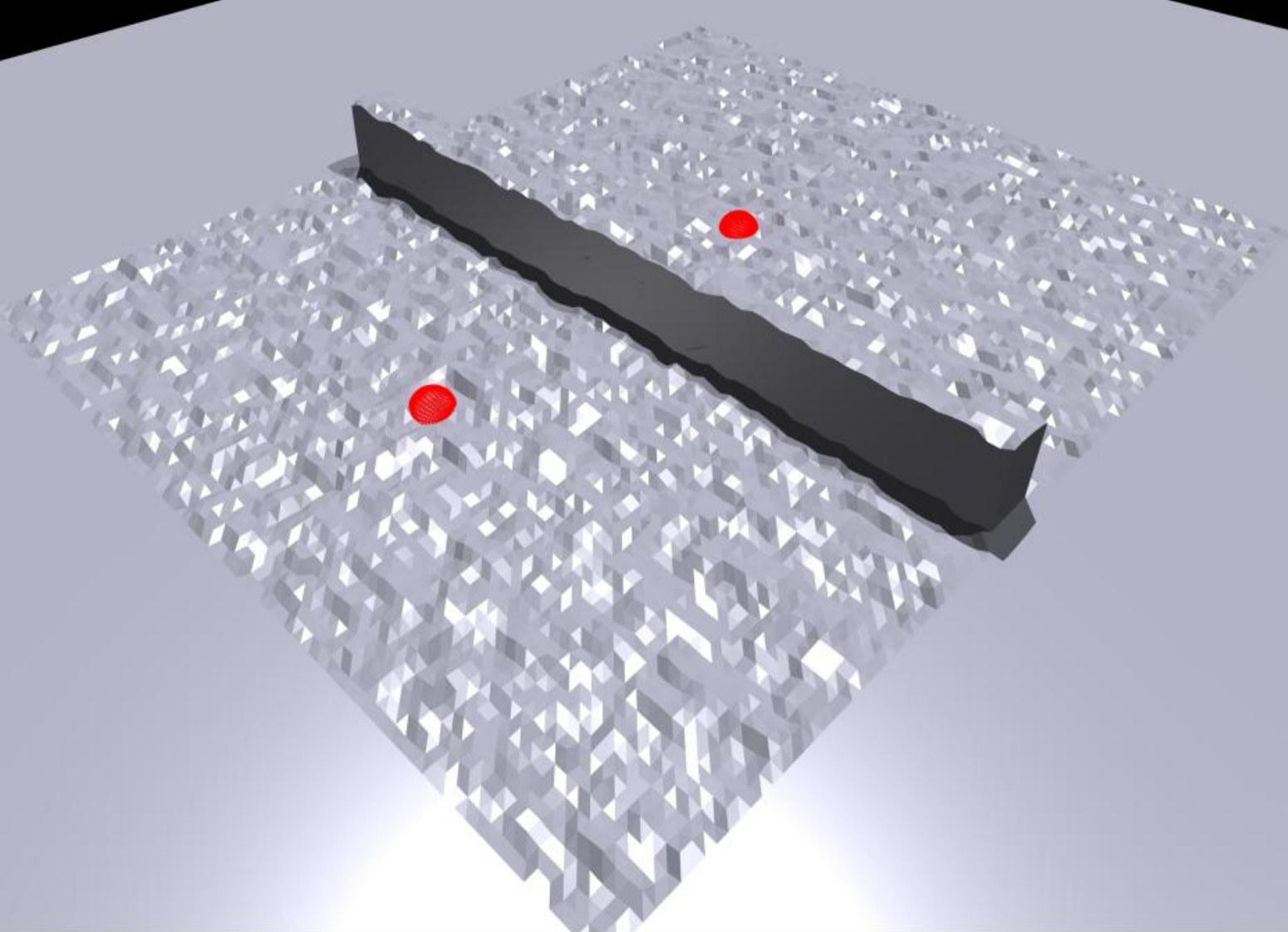}
        \subcaption{Hurdle}
        \label{hurdle}
    \end{minipage}
    \caption{
    Training scenario terrain units: In \emph{WP-Fixed}, units are aligned in a row; in \emph{WP-Random}, they are set in a matrix. Waypoints are preset only in the \emph{WP-Fixed} scenario.
    }
    \label{fig:terrain}
    \vspace{-5mm}
\end{wrapfigure}
Drawing inspiration from the agile performances of quadrupedal robots demonstrated in robot parkour studies \cite{extreme, robotparkour}, we select commonly used terrain types from their training setups, aiming for the locomotion policy to acquire diverse skills. For waypoint generation in this scenario, all waypoints are pre-set across the terrains based on the obstacle distribution, as illustrated in Fig. \ref{fig:terrain}.

Since the velocity command from the policy observation is replaced by a 2-dimensional waypoint, we have designed several reward functions tailored for waypoint tracking. To facilitate the robot's execution of a series of local navigation tasks, we introduce a reward function that encourages the robot to reach more waypoints:
\vspace{-0.15cm}
\begin{equation}
r_{\text{reach}} = n_p / ( t + \epsilon),
\vspace{-0.2cm}
\end{equation}
where $\epsilon$ represents a small constant, $n_p$ denotes the number of waypoints reached by the robot, and $t$ is the time elapsed since the beginning of the episode. To incentivize the robot to move towards the target waypoint, we design reward functions that consider the robot's yaw and velocity direction relative to the target waypoint position. Additionally, we have implemented a 'stay' reward function that allows the robot to remain at a waypoint after reaching it until the next waypoint is given:
\vspace{-0.15cm}
\begin{equation}
r_{\text{stay}} = \exp(-|\mathbf{q}_{\text{default}} - \mathbf{q}|) \mathbb{I}_{|\mathbf{w}_{\text{rel}}| < d_t},
\vspace{-0.2cm}
\end{equation}
where $\mathbf{q}_{\rm default}$ and $\mathbf{q}$ represent the default and current joint positions, respectively. $\mathbf{w}_{\rm rel}$ denotes the vector from the robot to the current target waypoint, and $\mathbb{I}$ is an indicator function that renders this reward non-zero only when the robot is near the target waypoint. During training, if the robot remains near a target waypoint for more than two seconds, the target shifts to the next preset waypoint. Consequently, during deployment, the robot can stay at the waypoint indefinitely while waiting for the next command. When $r_{\rm stay}$ is non-zero, all other reward functions are set to zero. Furthermore, other regularization terms are employed, as described in \cite{extreme, 2024HIM}.

\subsubsection{WP-Random scenario for ability enhancement} 

While the robot trained in the \emph{WP-Fixed} scenario acquires basic locomotion skills, it still struggles to continuously track waypoints that are randomly and irregularly distributed across complex terrain---an essential capability for a low-level controller in our navigation system. Consequently, we have established the \emph{WP-Random} scenario for fine-tuning the locomotion policy. 

This scenario is configured with a grid comprising \(n\) rows and \(m\) columns of terrain units, each having a fixed size.
We designate one row as flat terrain to serve as the robot's starting point for each episode. Unlike the \emph{WP-Fixed} scenario, where waypoints are predetermined, in the \emph{WP-Random} scenario, waypoints are dynamically selected based on the robot's position and yaw angle after it has remained at the last waypoint for two seconds. Waypoint candidates are positions within specific distance and orientation thresholds (i.e., \(\leq 90^\circ\)), where the distance threshold aligns with the size of the terrain units to prevent the robot from crossing multiple units. This setup of irregular and consecutive waypoints compels the robot to traverse diverse routes during training, thereby enhancing its ability to handle various situations.

\begin{wrapfigure}{r}{0.48\textwidth}
\centering
\includegraphics[width=0.48\textwidth]{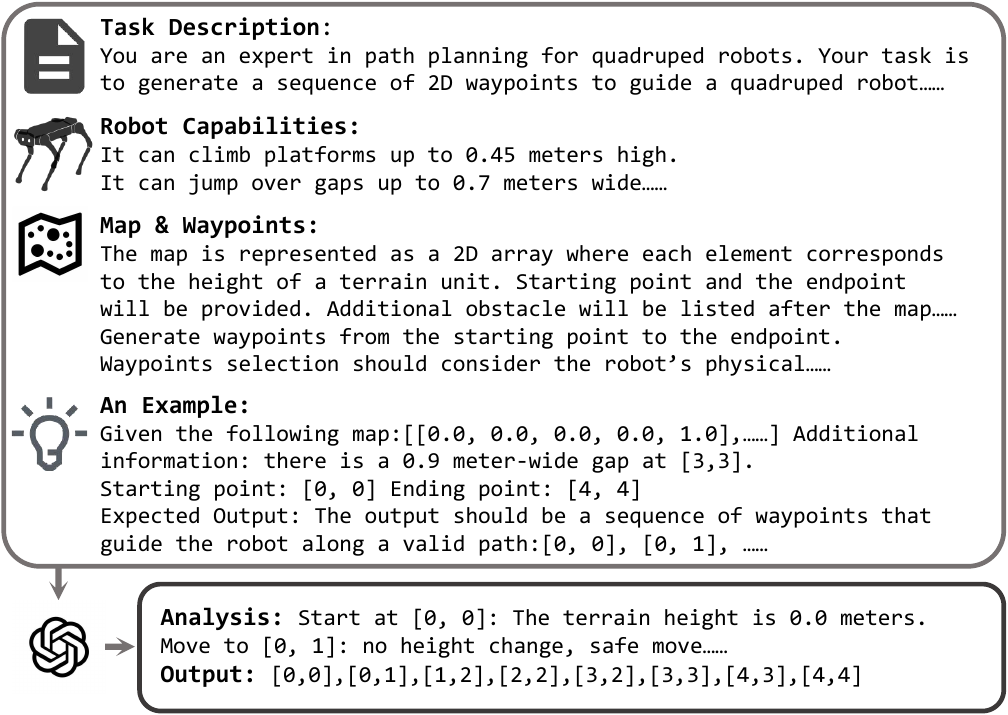}
\caption{Prompts for using the LLM to generate waypoints based on terrain information and the robot's capabilities.}
\vspace{-1em}
\vspace{-0.2cm}
\label{fig:prompt}
\end{wrapfigure}

To enhance obstacle avoidance, we randomly distribute variously sized obstacles within specific terrain units that the robot must bypass. Direct training on such terrains often leads to avoidable collisions and adjusting collision-related rewards to mitigate this issue proves to be quite tedious. To address this, we implement inflated virtual obstacles during policy distillation. By enlarging obstacle representations detected by the teacher policy in student-policy training without changing the actual obstacles, thus ensuring that the depth information remains accurately sized. This approach trains the student policy to maintain safe distances from obstacles, improving navigational safety.

Due to the high complexity of the \emph{WP-Random} scenario, directly fine-tuning the policy using the same reward functions as in the initial training phase could cause the robot to move slowly or even retreat when encountering obstacles or avoiding hazards. We modify the reward concerning velocity direction as follows:
\vspace{-0.15cm}
\begin{equation}
r_{\text{track}}=\left\{
\begin{array}{lcl}
-1,       &      & \mathbb{C}(\mathbf{v}, \mathbf{w}_{\text{rel}}) < 0.1\\
 \mathbb{C}(\mathbf{v},  \mathbf{w}_{\text{rel}}) |\mathbf{v}|,     &      & \text{others}
\end{array} \right. 
\vspace{-0.2cm}
\end{equation}
where \(\mathbb{C}(\cdot, \cdot)\) computes the cosine similarity between two vectors. In contrast to the initial training phase, we leverage the magnitude of velocity to encourage forward movement and penalize behaviors that significantly deviate from the direction oriented towards the waypoint. As the robot is required to navigate through complex terrains, we reduce the regularization terms related to vertical linear velocity and base orientation. Following \cite{robotparkour}, we also introduce noise to the depth image in simulations to minimize the discrepancy between simulated and real-world camera perceptions during policy distillation.

\subsection{Hierarchical Navigation Architecture}

To perform locomotion and navigation, previous work \cite{Anymalparkour, barkour} utilized fine-grained elevation maps. Generally, for navigation tasks, coarse-grained information like room layouts or approximate terrain heights is available. In Skill-Nav, the high-level planner generates waypoints using this coarse-grained information, without the need for high-precision sensors or prior knowledge. The limitations of coarse-grained maps are offset by versatile low-level policies that perform locomotion and obstacle avoidance. We have developed two types of high-level planners as follows.

\subsubsection{Classical Planning} Classical path planning algorithms such as \emph{A*} and \emph{Dijkstra} are employed to guide a RL-based quadrupedal locomotion controller \cite{wheeled-legged, IntentionNet}. We start by feeding an occupancy map annotated only with wall information and the start and end points into the high-level planner to derive a path. This path is then translated into a sequence of waypoints for the low-level controller.

\subsubsection{LLM Planner} 
\label{nav_method} 
Previous work has demonstrated that LLMs can perform embodied reasoning for various planning tasks with appropriate prompts \cite{palm_e}. To enable LLMs to act as a high-level planner in navigation, we prompt the LLM to generate waypoints that connect start and end positions given task descriptions, coarse-grained maps, low-level locomotion capabilities, terrain descriptions, and waypoint definitions. We give an example in Fig. \ref{fig:prompt}. The description of locomotion capabilities is based on the most difficult terrain the robot can traverse during the test. The LLM progressively reasons and produces a series of waypoints to direct the low-level controller.

\section{Experiments}

\subsection{Simulation Setup}

We utilize Isaac Gym to train our locomotion policy. Specifically, for the \emph{WP-Fixed} scenario, we create 10 rows and 40 columns of tracks, where each track contains 6 terrain units of the same type; and the \emph{WP-Random} scenario consists of 10 rows and 10 columns of areas, where each area containing 6×5 terrain units.  Table \ref{tab:terrain_property} provides the parameter ranges for critical obstacle properties such as box height and gap width. In the \emph{WP-Random} scenario, we apply constraints on the generation of waypoints and the arrangement of terrain units to ensure that each waypoint remains accessible to the robot. Notably, introducing noise during training plays a crucial role in enhancing the generalization ability of the policy \cite{li2022positive, zhang2025variational, huang2025enhance, zhang2024data, huang2025learn}. For sim-to-real transfer, we incorporate domain randomization and observation noise throughout the training process, similar to approaches used in other open-source works \cite{extreme, 2022leggedgym}. We train the teacher policy for 4000 iterations in the \emph{WP-Fixed} scenario and 8000 iterations in the \emph{WP-Random} scenario, followed by policy distillation to train the student policy for 6000 iterations. All training is conducted on an RTX-4090 GPU. We adopt Unitree AlienGo for simulation and real-world experiments.

\subsection{Hardware Deployment}
For real-world experiments, we deploy the Unitree Aliengo, equipped with a Jetson Orin NX for onboard computation and an Intel RealSense D435 for capturing depth images. We enhance and denoise images to align the depth data from simulations with real-world observations basically following \cite{extreme, robotparkour}. The robot features 12 degrees of freedom. Our control policy operates at 50Hz and utilizes a built-in PD controller at 200Hz with Kp = 40 and Kd = 2.0 for robot control. Additionally, a motion capture system is used to provide localization and ensure stable deployment performance.

\begin{table}[t]
\centering 
\caption{
Parameter ranges of key obstacle properties for the terrain curriculum, measured in meters.}
\label{tab:terrain_property}
\fontsize{10}{12}\selectfont
\begin{tabular}{cccc}
\toprule
\multirow{2}{*}{Terrain Type} & \multirow{2}{*}{Property} & WP-Fixed   & WP-Random \\
  & & Scenario & Scenario \\
\hline
Hurdle     &  height &  (0.1, 0.4) &  (0.1, 0.3)\\
Box      &  height &  (0.1, 0.5) &  (0.1, 0.35)\\ 
Gap  &  width & (0.1, 0.9) & (0.1, 0.35)\\
Obstacle   &  size & (0.15, 1.0)  & (0.2, 0.9)\\ 
\bottomrule
\end{tabular}
\end{table}

\begin{table}[t]
\centering
\caption{
Results on test tasks. Average Travel Distance (ATD) and Average Success Time (AST) are reported in meters and seconds, respectively. SR denotes Success Rate, while ‘obst.’ refers to high obstacles. Ext. Par. stands for \textbf{Extreme Parkour}. The '/'  indicates no successful episodes.}
\label{tab:sim_result}
\fontsize{10}{12}\selectfont
\begin{tabular}{p{1.cm}p{1.5cm}|p{0.5cm}p{0.7cm}p{0.7cm}|p{0.5cm}p{0.7cm}p{0.65cm}}
\toprule
\multirow{2}{*}{\makecell{Test \\ Task}} & \multirow{2}{*}{Method} & \multicolumn{3}{c|}{w/o obst.} & \multicolumn{3}{c}{with obst.} \\ & & SR$\uparrow$ & ATD$\uparrow$ & AST$\downarrow$ & SR$\uparrow$ & ATD$\uparrow$ & AST$\downarrow$ \\ \hline
\multirow{5}{*}{Single} & RMA & 0.00 & 3.3 & / & 0.00 & 4.7 & / \\
& Ext. Par. & 1.00 & 15.8 & \textbf{15.2} & 0.00 & 9.9 & / \\
& Ours-s1 & 0.83 & 13.7 & 23.0 & 0.83 & 14.2 & 22.8 \\
& Ours-s2 & 1.00 & 15.8 & 17.2 & 0.00 & 12.0 & / \\
& Ours & \textbf{1.00} & \textbf{15.8} & 17.6 & \textbf{1.00} & \textbf{15.8} & \textbf{18.3} \\ \hline
\multirow{5}{*}{Omni} & RMA & 0.00 & 5.0 & / & 0.00 & 2.6 & / \\
& Ext. Par. & 0.44 & 5.6 & 12.9 & 0.28 & 5.3 & 14.0 \\
& Ours-s1 & 0.17 & 4.6 & 15.7 & 0.22 & 4.4 & 15.2 \\
& Ours-s2 & 0.89 & 7.9 & 10.3 & 0.83 & 8.1 & 11.2 \\
& Ours & \textbf{0.89} & \textbf{8.2} & \textbf{9.9} & \textbf{0.89} & \textbf{8.2} & \textbf{11.0} \\
\bottomrule
\end{tabular}
\vspace{-0.2cm}
\end{table}

\subsection{Locomotion Comparison}
\label{loco-comp}
To analyze the capabilities of our low-level locomotion policy and the importance of the two training scenarios, we compare our locomotion policy with the following baselines and ablations:
\begin{itemize}
    \item \textbf{RMA}: Training the Rapid Motor Adaptation\cite{2021rma} using privileged information including terrain scandots with terrain setups referring to \cite{2022leggedgym}.
    \item \textbf{Extreme Parkour}: Designing reward functions using waypoints to guide a robot to perform parkour on extreme terrains with a teacher-student architecture\cite{extreme}.
    \item \textbf{Ours-s1}: Training a teacher policy only in \emph{WP-Fixed} scenario without fine-tuning, and then distilling the teacher policy into a student policy.
    \item \textbf{Ours-s2}: Training a teacher policy directly in \emph{WP-Random} scenario without pre-training, and distilling the teacher policy into a student policy.
\end{itemize}

\begin{figure*}[t]
\centering
\includegraphics[width=\textwidth]{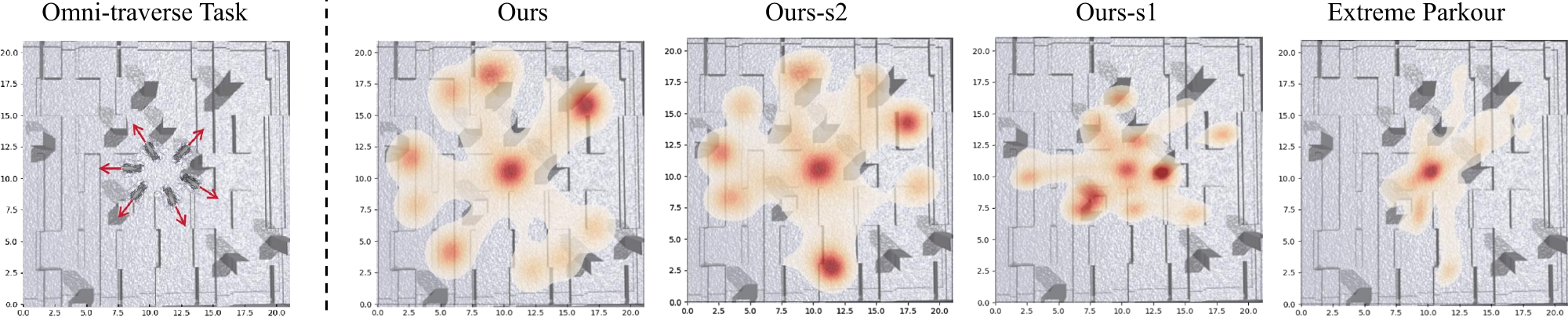}
\caption{
Demonstration of the \emph{omni-traverse} task (\textit{left}), where robots move from the center toward the boundary. The heatmaps (\textit{right}) shows task methods execution; darker colors indicate more visits, and larger areas reflect better terrain traversal.
}
\vspace{-1em}
\vspace{-0.2cm}
\label{fig:heatmaps}
\end{figure*}

The comparison between our method and those of \emph{RMA} and \emph{Extreme Parkour} aims to validate the locomotion capabilities based on waypoint commands. Additionally, contrasting our method with two ablations underscores the significance of utilizing dual training scenarios. For evaluation, we designed two test tasks and deployed 18 robots for each task. In the \emph{single-traverse} task, all robots are required to move in the same direction and traverse a series of obstacles. The \emph{omni-traverse} task, depicted in Fig. \ref{fig:heatmaps} (\textit{left}), positions robots at the center of a 21×21m terrain with varying fixed orientations. They then move outward from the center according to instructions (waypoints for our policy and velocity vectors for others). A robot is reinitialized to the starting point upon failure, such as collision or falling; if the task is completed successfully, the robot remains stationary. Three metrics are used to evaluate robot performance: Success Rate (SR), Average Travel Distance (ATD) from the origin, and Average Success Time (AST). In the \emph{single-traverse} task, successfully traversing the entire terrain within 30 seconds is deemed a success. In the \emph{omni-traverse} task, moving more than 8.5 meters from the center within 16 seconds is considered successful. For AST, episodes that do not result in success are recorded as the test maximum time if SR is non-zero. The obstacle parameters in the test terrains are identical to those at the highest curriculum level of the \emph{WP-Random} scenario. 

We conducted experiments with and without high obstacles for both test tasks. Results presented in Table \ref{tab:sim_result} reveal that RMA is unable to traverse the terrain, and robust walking without locomotion skills is insufficient to address these tasks. In the \emph{single-traverse} task without high obstacles, all methods performed well; \textbf{Extreme Parkour} completed the task the fastest due to specific training in similar terrains, despite suffering from significant tracking errors. The failure of \textbf{Ours-s1} can be attributed to the limited and regular distribution of waypoints provided during training. Conversely, \textbf{Ours-s2} fails because training directly in the \emph{WP-Random} scenario led to a terrible gait characterized by excessive jumping, rendering it unsuitable for deployment. Additionally, \textbf{Ours-s2} struggles to bypass wide high obstacles directly in its path. Despite the presence of high obstacles, our method outperforms the others in the \emph{omni-traverse} task. This superiority is due to the policy being trained in both scenarios, enabling it to handle more irregular local navigation situations. In some instances, \textbf{Extreme Parkour} deviates from the commands and approaches obstacles that could have been avoided. Heatmaps in Fig.~\ref{fig:heatmaps} illustrate the position visiting frequencies of different methods in the \emph{omni-traverse} task. Both \textbf{Ours-s1} and \textbf{Extreme Parkour} struggle to navigate irregular obstacles, whereas our method visits farther positions more frequently, demonstrating more powerful locomotion skills. In the \emph{omni-traverse} task with complex terrains and high obstacles, although our method may encounter collisions at terrain edges, leading to a certain probability of failure, it can still successfully accomplish the task in most cases.

\begin{figure}[t]
\centering
\includegraphics[width=0.49\textwidth]{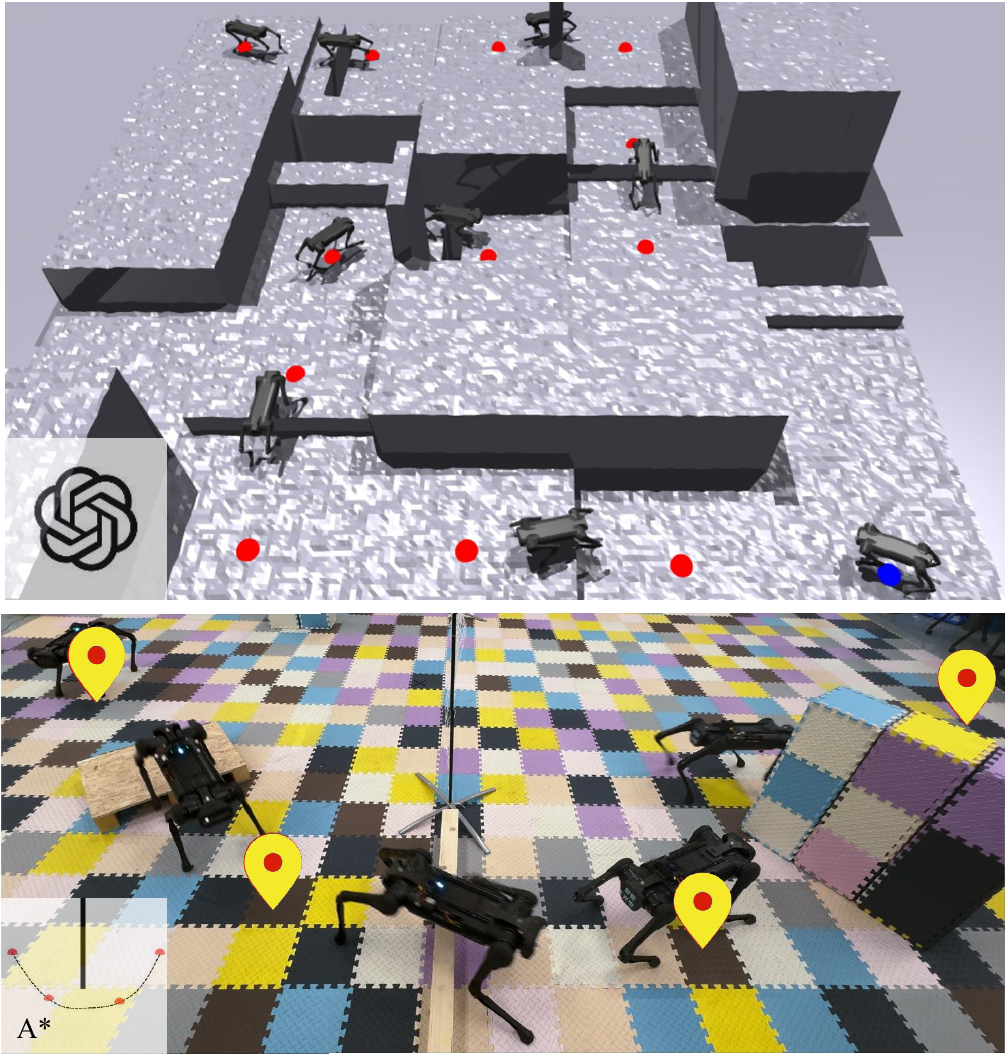}
\caption{Snapshots of the robot navigating in both simulated and real-world environments. The bottom-left icons indicate waypoints planning using GPT and A* algorithms.}
\vspace{-1em}
\label{fig:llmnav}
\end{figure}

\subsection{Navigation System}

Fig. \ref{fig:llmnav} showcases sequential snapshots of our system navigating in a simulation using GPT-4\cite{achiam2023gpt}, and in a real-world scenario using the \emph{A*} algorithm. In the simulation, the navigation terrain consists of multiple terrain units, each 2m $\times$ 2m, featuring varied heights and obstacles. The robot is initialized in one terrain unit, with another designated as the navigation target. The LLM (specifically, GPT-4) receives prompts as described in \ref{nav_method} and outputs waypoints in the form of terrain unit indices. These waypoints, positioned at the centers of the corresponding terrain units, guide our low-level controller to reach the target while handle obstacles not accounted for by the LLM. However, this waypoint configuration often overlooks intricate terrain details, potentially leading to impractical waypoint placements, such as within high obstacles. Our low-level controller can struggle to recover from some of these bad cases such as a waypoint in the middle of the gap or at the edge of a box, even though it has not been trained for such anomalies. Specifically, if the waypoint to be tracked by the robot is located in a gap region, the robot may experience partial leg suspension due to the anomalous waypoint. To maintain balance, the robot will either straddle the center of the gap or position itself close to the gap edge, rather than falling directly into the gap and causing termination.

In our real-world experiments, we employ a high-frequency motion capture system for localization. As shown in Fig. \ref{fig:llmnav}, the navigation environment is rendered into an occupancy map that marks only wall positions. Applying the \emph{A*} algorithm on this map, we generate paths as a series of continuous coordinate points.
These points are uniformly segmented into waypoints at intervals ranging from 0.5 to 3 meters, which are subsequently provided as inputs to the low-level controller. The robot successfully navigates to the target waypoint, adeptly handling obstacles. Even when the robot encounters low obstacles that the depth camera fails to detect, it is capable of regaining balance and continuing to the target. Furthermore, despite external forces that may divert the robot from its intended path, it manages to complete its task. If the waypoint's direction significantly deviates from the robot's current orientation, the robot adjusts by turning quickly to align with the waypoint.

\section{Conclusion}
In this work, we introduced the use of waypoints as an interface to connect a versatile quadrupedal robot with a high-level planner for navigation. The locomotion policy trained in two simulated scenarios is able to traverse complex terrains while avoiding obstacles, and the high-level planner such as LLMs enhances generalization across various environments. Experiments demonstrate that the proposed navigation architecture is able to complete navigation tasks in complex scenarios. Future works include designing an edge collision-free low-level controller for more robust locomotion and  developing an end-to-end policy to conduct locomotion and navigation simultaneously.
\backmatter

\section*{\small Abbreviations}
\begin{tabular}{ll}
    LLM      & Large language models \\
    MPC      & Model predictive control \\
    RL       & Reinforcement learning \\
    RNN      & Recurrent neural network \\ 
    PPO      & Proximal policy optimization \\
    MDP      & Markov decision process \\
    \emph{WP-Fixed} & Fixed waypoints \\
    \emph{WP-Random} & Random waypoints \\
    PD       & Proportional-Derivative \\
    SR       & Success rate \\
    ATD      & Average travel distance \\
    AST      & Average success time
\end{tabular}

\section*{\small Code availability}
The codes of baseline methods are from the open-source works. The codes of Skill-Nav are available from the first author.

\section*{\normalsize Declarations}

\section*{\small Availability of data and material}
All data generated or analyzed during this study are included in this manuscript.

\section*{\small Competing interests}
The authors declare that they have no known competing financial interests or personal relationships that could have appeared to influence the work reported in this paper, may be limited by the attached paper reports the research.

\section*{\small Funding}
This work is supported by the National Natural Science Foundation of China (Grant No.62306242),  the Young Elite Scientists Sponsorship Program by CAST (Grant No. 2024QNRC001), and the Yangfan Project of the Shanghai (Grant No.23YF11462200).

\section*{\small Authors' contributions}
All authors contributed to the study conception and design. Dewei Wang established the overall framework of the study and drafted the manuscript. Chenjia Bai improved the manuscript and guided the progress of the work. Chenhui Li and Jiyuan Shi participated in the revision of the manuscript and the implementation of the experiments. Yan Ding, Chi Zhang and Bin Zhao guided the research direction and the execution of the experiments.

\section*{\small Acknowledgements}
We sincerely thank Dr. Minghuan Liu for his guidance and support in both the writing of this paper and the hardware deployment experiments.

\bibliography{sn-bibliography}


\begin{thebibliography}{67}
\ifx \bisbn   \undefined \def \bisbn  #1{ISBN #1}\fi
\ifx \binits  \undefined \def \binits#1{#1}\fi
\ifx \bauthor  \undefined \def \bauthor#1{#1}\fi
\ifx \batitle  \undefined \def \batitle#1{#1}\fi
\ifx \bjtitle  \undefined \def \bjtitle#1{#1}\fi
\ifx \bvolume  \undefined \def \bvolume#1{\textbf{#1}}\fi
\ifx \byear  \undefined \def \byear#1{#1}\fi
\ifx \bissue  \undefined \def \bissue#1{#1}\fi
\ifx \bfpage  \undefined \def \bfpage#1{#1}\fi
\ifx \blpage  \undefined \def \blpage #1{#1}\fi
\ifx \burl  \undefined \def \burl#1{\textsf{#1}}\fi
\ifx \doiurl  \undefined \def \doiurl#1{\url{https://doi.org/#1}}\fi
\ifx \betal  \undefined \def \betal{\textit{et al.}}\fi
\ifx \binstitute  \undefined \def \binstitute#1{#1}\fi
\ifx \binstitutionaled  \undefined \def \binstitutionaled#1{#1}\fi
\ifx \bctitle  \undefined \def \bctitle#1{#1}\fi
\ifx \beditor  \undefined \def \beditor#1{#1}\fi
\ifx \bpublisher  \undefined \def \bpublisher#1{#1}\fi
\ifx \bbtitle  \undefined \def \bbtitle#1{#1}\fi
\ifx \bedition  \undefined \def \bedition#1{#1}\fi
\ifx \bseriesno  \undefined \def \bseriesno#1{#1}\fi
\ifx \blocation  \undefined \def \blocation#1{#1}\fi
\ifx \bsertitle  \undefined \def \bsertitle#1{#1}\fi
\ifx \bsnm \undefined \def \bsnm#1{#1}\fi
\ifx \bsuffix \undefined \def \bsuffix#1{#1}\fi
\ifx \bparticle \undefined \def \bparticle#1{#1}\fi
\ifx \barticle \undefined \def \barticle#1{#1}\fi
\bibcommenthead
\ifx \bconfdate \undefined \def \bconfdate #1{#1}\fi
\ifx \botherref \undefined \def \botherref #1{#1}\fi
\ifx \url \undefined \def \url#1{\textsf{#1}}\fi
\ifx \bchapter \undefined \def \bchapter#1{#1}\fi
\ifx \bbook \undefined \def \bbook#1{#1}\fi
\ifx \bcomment \undefined \def \bcomment#1{#1}\fi
\ifx \oauthor \undefined \def \oauthor#1{#1}\fi
\ifx \citeauthoryear \undefined \def \citeauthoryear#1{#1}\fi
\ifx \endbibitem  \undefined \def \endbibitem {}\fi
\ifx \bconflocation  \undefined \def \bconflocation#1{#1}\fi
\ifx \arxivurl  \undefined \def \arxivurl#1{\textsf{#1}}\fi
\csname PreBibitemsHook\endcsname

\bibitem[\protect\citeauthoryear{Jiang et~al.}{2018}]{jiang2018deep}
\begin{barticle}
\bauthor{\bsnm{Jiang}, \binits{X.}},
\bauthor{\bsnm{Pang}, \binits{Y.}},
\bauthor{\bsnm{Li}, \binits{X.}},
\bauthor{\bsnm{Pan}, \binits{J.}},
\bauthor{\bsnm{Xie}, \binits{Y.}}:
\batitle{Deep neural networks with elastic rectified linear units for object recognition}.
\bjtitle{Neurocomputing}
\bvolume{275},
\bfpage{1132}--\blpage{1139}
(\byear{2018})
\end{barticle}
\endbibitem

\bibitem[\protect\citeauthoryear{Gao et~al.}{2015}]{gao2015learning}
\begin{barticle}
\bauthor{\bsnm{Gao}, \binits{F.}},
\bauthor{\bsnm{Tao}, \binits{D.}},
\bauthor{\bsnm{Gao}, \binits{X.}},
\bauthor{\bsnm{Li}, \binits{X.}}:
\batitle{Learning to rank for blind image quality assessment}.
\bjtitle{IEEE transactions on neural networks and learning systems}
\bvolume{26}(\bissue{10}),
\bfpage{2275}--\blpage{2290}
(\byear{2015})
\end{barticle}
\endbibitem

\bibitem[\protect\citeauthoryear{Tao et~al.}{2006}]{tao2006human}
\begin{bchapter}
\bauthor{\bsnm{Tao}, \binits{D.}},
\bauthor{\bsnm{Li}, \binits{X.}},
\bauthor{\bsnm{Maybank}, \binits{S.J.}},
\bauthor{\bsnm{Wu}, \binits{X.}}:
\bctitle{Human carrying status in visual surveillance}.
In: \bbtitle{2006 IEEE Computer Society Conference on Computer Vision and Pattern Recognition (CVPR'06)},
vol. \bseriesno{2},
pp. \bfpage{1670}--\blpage{1677}
(\byear{2006}).
\bcomment{IEEE}
\end{bchapter}
\endbibitem

\bibitem[\protect\citeauthoryear{Han et~al.}{2015}]{han2015two}
\begin{barticle}
\bauthor{\bsnm{Han}, \binits{J.}},
\bauthor{\bsnm{Zhang}, \binits{D.}},
\bauthor{\bsnm{Wen}, \binits{S.}},
\bauthor{\bsnm{Guo}, \binits{L.}},
\bauthor{\bsnm{Liu}, \binits{T.}},
\bauthor{\bsnm{Li}, \binits{X.}}:
\batitle{Two-stage learning to predict human eye fixations via sdaes}.
\bjtitle{IEEE transactions on cybernetics}
\bvolume{46}(\bissue{2}),
\bfpage{487}--\blpage{498}
(\byear{2015})
\end{barticle}
\endbibitem

\bibitem[\protect\citeauthoryear{Tao et~al.}{2008}]{tao2008bayesian}
\begin{barticle}
\bauthor{\bsnm{Tao}, \binits{D.}},
\bauthor{\bsnm{Song}, \binits{M.}},
\bauthor{\bsnm{Li}, \binits{X.}},
\bauthor{\bsnm{Shen}, \binits{J.}},
\bauthor{\bsnm{Sun}, \binits{J.}},
\bauthor{\bsnm{Wu}, \binits{X.}},
\bauthor{\bsnm{Faloutsos}, \binits{C.}},
\bauthor{\bsnm{Maybank}, \binits{S.J.}}:
\batitle{Bayesian tensor approach for 3-d face modeling}.
\bjtitle{IEEE Transactions on Circuits and Systems for Video Technology}
\bvolume{18}(\bissue{10}),
\bfpage{1397}--\blpage{1410}
(\byear{2008})
\end{barticle}
\endbibitem

\bibitem[\protect\citeauthoryear{Arm et~al.}{2023}]{arm2023scientific}
\begin{barticle}
\bauthor{\bsnm{Arm}, \binits{P.}},
\bauthor{\bsnm{Waibel}, \binits{G.}},
\bauthor{\bsnm{Preisig}, \binits{J.}},
\bauthor{\bsnm{Tuna}, \binits{T.}},
\bauthor{\bsnm{Zhou}, \binits{R.}},
\bauthor{\bsnm{Bickel}, \binits{V.}},
\bauthor{\bsnm{Ligeza}, \binits{G.}},
\bauthor{\bsnm{Miki}, \binits{T.}},
\bauthor{\bsnm{Kehl}, \binits{F.}},
\bauthor{\bsnm{Kolvenbach}, \binits{H.}}, \betal:
\batitle{Scientific exploration of challenging planetary analog environments with a team of legged robots}.
\bjtitle{Science robotics}
\bvolume{8}(\bissue{80}),
\bfpage{9548}
(\byear{2023})
\end{barticle}
\endbibitem

\bibitem[\protect\citeauthoryear{Tranzatto et~al.}{2022}]{2022cerberus}
\begin{barticle}
\bauthor{\bsnm{Tranzatto}, \binits{M.}},
\bauthor{\bsnm{Miki}, \binits{T.}},
\bauthor{\bsnm{Dharmadhikari}, \binits{M.}},
\bauthor{\bsnm{Bernreiter}, \binits{L.}},
\bauthor{\bsnm{Kulkarni}, \binits{M.}},
\bauthor{\bsnm{Mascarich}, \binits{F.}},
\bauthor{\bsnm{Andersson}, \binits{O.}},
\bauthor{\bsnm{Khattak}, \binits{S.}},
\bauthor{\bsnm{Hutter}, \binits{M.}},
\bauthor{\bsnm{Siegwart}, \binits{R.}}, \betal:
\batitle{Cerberus in the darpa subterranean challenge}.
\bjtitle{Science Robotics}
\bvolume{7}(\bissue{66}),
\bfpage{9742}
(\byear{2022})
\end{barticle}
\endbibitem

\bibitem[\protect\citeauthoryear{Qu et~al.}{2025}]{qu2025spatialvla}
\begin{botherref}
\oauthor{\bsnm{Qu}, \binits{D.}},
\oauthor{\bsnm{Song}, \binits{H.}},
\oauthor{\bsnm{Chen}, \binits{Q.}},
\oauthor{\bsnm{Yao}, \binits{Y.}},
\oauthor{\bsnm{Ye}, \binits{X.}},
\oauthor{\bsnm{Ding}, \binits{Y.}},
\oauthor{\bsnm{Wang}, \binits{Z.}},
\oauthor{\bsnm{Gu}, \binits{J.}},
\oauthor{\bsnm{Zhao}, \binits{B.}},
\oauthor{\bsnm{Wang}, \binits{D.}}, et al.:
Spatialvla: Exploring spatial representations for visual-language-action model.
arXiv preprint arXiv:2501.15830
(2025)
\end{botherref}
\endbibitem

\bibitem[\protect\citeauthoryear{Song et~al.}{2025}]{song2025hume}
\begin{botherref}
\oauthor{\bsnm{Song}, \binits{H.}},
\oauthor{\bsnm{Qu}, \binits{D.}},
\oauthor{\bsnm{Yao}, \binits{Y.}},
\oauthor{\bsnm{Chen}, \binits{Q.}},
\oauthor{\bsnm{Lv}, \binits{Q.}},
\oauthor{\bsnm{Tang}, \binits{Y.}},
\oauthor{\bsnm{Shi}, \binits{M.}},
\oauthor{\bsnm{Ren}, \binits{G.}},
\oauthor{\bsnm{Yao}, \binits{M.}},
\oauthor{\bsnm{Zhao}, \binits{B.}}, et al.:
Hume: Introducing system-2 thinking in visual-language-action model.
arXiv preprint arXiv:2505.21432
(2025)
\end{botherref}
\endbibitem

\bibitem[\protect\citeauthoryear{An et~al.}{2025}]{an2025aiflowperspectivesscenarios}
\begin{botherref}
\oauthor{\bsnm{An}, \binits{H.}},
\oauthor{\bsnm{Hu}, \binits{W.}},
\oauthor{\bsnm{Huang}, \binits{S.}},
\oauthor{\bsnm{Huang}, \binits{S.}},
\oauthor{\bsnm{Li}, \binits{R.}},
\oauthor{\bsnm{Liang}, \binits{Y.}},
\oauthor{\bsnm{Shao}, \binits{J.}},
\oauthor{\bsnm{Song}, \binits{Y.}},
\oauthor{\bsnm{Wang}, \binits{Z.}},
\oauthor{\bsnm{Yuan}, \binits{C.}},
\oauthor{\bsnm{Zhang}, \binits{C.}},
\oauthor{\bsnm{Zhang}, \binits{H.}},
\oauthor{\bsnm{Zhuang}, \binits{W.}},
\oauthor{\bsnm{Li}, \binits{X.}}:
AI Flow: Perspectives, Scenarios, and Approaches
(2025).
\url{https://arxiv.org/abs/2506.12479}
\end{botherref}
\endbibitem

\bibitem[\protect\citeauthoryear{Wang et~al.}{2025}]{wang2025more}
\begin{botherref}
\oauthor{\bsnm{Wang}, \binits{D.}},
\oauthor{\bsnm{Wang}, \binits{X.}},
\oauthor{\bsnm{Liu}, \binits{X.}},
\oauthor{\bsnm{Shi}, \binits{J.}},
\oauthor{\bsnm{Zhao}, \binits{Y.}},
\oauthor{\bsnm{Bai}, \binits{C.}},
\oauthor{\bsnm{Li}, \binits{X.}}:
More: Mixture of residual experts for humanoid lifelike gaits learning on complex terrains.
arXiv preprint arXiv:2506.08840
(2025)
\end{botherref}
\endbibitem

\bibitem[\protect\citeauthoryear{Winkler et~al.}{2018}]{2018MPC2}
\begin{barticle}
\bauthor{\bsnm{Winkler}, \binits{A.W.}},
\bauthor{\bsnm{Bellicoso}, \binits{C.D.}},
\bauthor{\bsnm{Hutter}, \binits{M.}},
\bauthor{\bsnm{Buchli}, \binits{J.}}:
\batitle{Gait and trajectory optimization for legged systems through phase-based end-effector parameterization}.
\bjtitle{IEEE Robotics and Automation Letters}
\bvolume{3}(\bissue{3}),
\bfpage{1560}--\blpage{1567}
(\byear{2018})
\end{barticle}
\endbibitem

\bibitem[\protect\citeauthoryear{Lee et~al.}{2020}]{2020challengingterrain}
\begin{barticle}
\bauthor{\bsnm{Lee}, \binits{J.}},
\bauthor{\bsnm{Hwangbo}, \binits{J.}},
\bauthor{\bsnm{Wellhausen}, \binits{L.}},
\bauthor{\bsnm{Koltun}, \binits{V.}},
\bauthor{\bsnm{Hutter}, \binits{M.}}:
\batitle{Learning quadrupedal locomotion over challenging terrain}.
\bjtitle{Science robotics}
\bvolume{5}(\bissue{47}),
\bfpage{5986}
(\byear{2020})
\end{barticle}
\endbibitem

\bibitem[\protect\citeauthoryear{Kumar et~al.}{2021}]{2021rma}
\begin{bchapter}
\bauthor{\bsnm{Kumar}, \binits{A.}},
\bauthor{\bsnm{Fu}, \binits{Z.}},
\bauthor{\bsnm{Pathak}, \binits{D.}},
\bauthor{\bsnm{Malik}, \binits{J.}}:
\bctitle{Rma: Rapid motor adaptation for legged robots}.
In: \bbtitle{Robotics: Science and Systems}
(\byear{2021})
\end{bchapter}
\endbibitem

\bibitem[\protect\citeauthoryear{Miki et~al.}{2022}]{2022inthewild}
\begin{barticle}
\bauthor{\bsnm{Miki}, \binits{T.}},
\bauthor{\bsnm{Lee}, \binits{J.}},
\bauthor{\bsnm{Hwangbo}, \binits{J.}},
\bauthor{\bsnm{Wellhausen}, \binits{L.}},
\bauthor{\bsnm{Koltun}, \binits{V.}},
\bauthor{\bsnm{Hutter}, \binits{M.}}:
\batitle{Learning robust perceptive locomotion for quadrupedal robots in the wild}.
\bjtitle{Science robotics}
\bvolume{7}(\bissue{62}),
\bfpage{2822}
(\byear{2022})
\end{barticle}
\endbibitem

\bibitem[\protect\citeauthoryear{Rudin et~al.}{2022}]{2022leggedgym}
\begin{bchapter}
\bauthor{\bsnm{Rudin}, \binits{N.}},
\bauthor{\bsnm{Hoeller}, \binits{D.}},
\bauthor{\bsnm{Reist}, \binits{P.}},
\bauthor{\bsnm{Hutter}, \binits{M.}}:
\bctitle{Learning to walk in minutes using massively parallel deep reinforcement learning}.
In: \bbtitle{Conference on Robot Learning},
pp. \bfpage{91}--\blpage{100}
(\byear{2022}).
\bcomment{PMLR}
\end{bchapter}
\endbibitem

\bibitem[\protect\citeauthoryear{Makoviychuk et~al.}{2021}]{isaacgym}
\begin{botherref}
\oauthor{\bsnm{Makoviychuk}, \binits{V.}},
\oauthor{\bsnm{Wawrzyniak}, \binits{L.}},
\oauthor{\bsnm{Guo}, \binits{Y.}},
\oauthor{\bsnm{Lu}, \binits{M.}},
\oauthor{\bsnm{Storey}, \binits{K.}},
\oauthor{\bsnm{Macklin}, \binits{M.}},
\oauthor{\bsnm{Hoeller}, \binits{D.}},
\oauthor{\bsnm{Rudin}, \binits{N.}},
\oauthor{\bsnm{Allshire}, \binits{A.}},
\oauthor{\bsnm{Handa}, \binits{A.}}, et al.:
Isaac gym: High performance gpu-based physics simulation for robot learning.
arXiv preprint arXiv:2108.10470
(2021)
\end{botherref}
\endbibitem

\bibitem[\protect\citeauthoryear{Bengio et~al.}{2009}]{bengio2009curriculum}
\begin{bchapter}
\bauthor{\bsnm{Bengio}, \binits{Y.}},
\bauthor{\bsnm{Louradour}, \binits{J.}},
\bauthor{\bsnm{Collobert}, \binits{R.}},
\bauthor{\bsnm{Weston}, \binits{J.}}:
\bctitle{Curriculum learning}.
In: \bbtitle{Proceedings of the 26th Annual International Conference on Machine Learning},
pp. \bfpage{41}--\blpage{48}
(\byear{2009})
\end{bchapter}
\endbibitem

\bibitem[\protect\citeauthoryear{Roth et~al.}{2024}]{2024viplanner}
\begin{bchapter}
\bauthor{\bsnm{Roth}, \binits{P.}},
\bauthor{\bsnm{Nubert}, \binits{J.}},
\bauthor{\bsnm{Yang}, \binits{F.}},
\bauthor{\bsnm{Mittal}, \binits{M.}},
\bauthor{\bsnm{Hutter}, \binits{M.}}:
\bctitle{Viplanner: Visual semantic imperative learning for local navigation}.
In: \bbtitle{2024 IEEE International Conference on Robotics and Automation (ICRA)},
pp. \bfpage{5243}--\blpage{5249}
(\byear{2024}).
\bcomment{IEEE}
\end{bchapter}
\endbibitem

\bibitem[\protect\citeauthoryear{Kareer et~al.}{2023}]{vinl}
\begin{bchapter}
\bauthor{\bsnm{Kareer}, \binits{S.}},
\bauthor{\bsnm{Yokoyama}, \binits{N.}},
\bauthor{\bsnm{Batra}, \binits{D.}},
\bauthor{\bsnm{Ha}, \binits{S.}},
\bauthor{\bsnm{Truong}, \binits{J.}}:
\bctitle{Vinl: Visual navigation and locomotion over obstacles}.
In: \bbtitle{2023 IEEE International Conference on Robotics and Automation (ICRA)},
pp. \bfpage{2018}--\blpage{2024}
(\byear{2023}).
\bcomment{IEEE}
\end{bchapter}
\endbibitem

\bibitem[\protect\citeauthoryear{Mattamala et~al.}{2022}]{efficientsafenav}
\begin{barticle}
\bauthor{\bsnm{Mattamala}, \binits{M.}},
\bauthor{\bsnm{Chebrolu}, \binits{N.}},
\bauthor{\bsnm{Fallon}, \binits{M.}}:
\batitle{An efficient locally reactive controller for safe navigation in visual teach and repeat missions}.
\bjtitle{IEEE Robotics and Automation Letters}
\bvolume{7}(\bissue{2}),
\bfpage{2353}--\blpage{2360}
(\byear{2022})
\end{barticle}
\endbibitem

\bibitem[\protect\citeauthoryear{Caluwaerts et~al.}{2023}]{barkour}
\begin{botherref}
\oauthor{\bsnm{Caluwaerts}, \binits{K.}},
\oauthor{\bsnm{Iscen}, \binits{A.}},
\oauthor{\bsnm{Kew}, \binits{J.C.}},
\oauthor{\bsnm{Yu}, \binits{W.}},
\oauthor{\bsnm{Zhang}, \binits{T.}},
\oauthor{\bsnm{Freeman}, \binits{D.}},
\oauthor{\bsnm{Lee}, \binits{K.-H.}},
\oauthor{\bsnm{Lee}, \binits{L.}},
\oauthor{\bsnm{Saliceti}, \binits{S.}},
\oauthor{\bsnm{Zhuang}, \binits{V.}}, et al.:
Barkour: Benchmarking animal-level agility with quadruped robots.
arXiv preprint arXiv:2305.14654
(2023)
\end{botherref}
\endbibitem

\bibitem[\protect\citeauthoryear{Xu et~al.}{2024}]{xu2024optimal}
\begin{bchapter}
\bauthor{\bsnm{Xu}, \binits{S.}},
\bauthor{\bsnm{Zhang}, \binits{W.}},
\bauthor{\bsnm{Ho}, \binits{C.P.}},
\bauthor{\bsnm{Zhu}, \binits{L.}}:
\bctitle{Optimal prescribed-time control based reactive planning system for quadruped robot navigation}.
In: \bbtitle{2024 IEEE International Conference on Robotics and Automation (ICRA)},
pp. \bfpage{13185}--\blpage{13191}
(\byear{2024}).
\bcomment{IEEE}
\end{bchapter}
\endbibitem

\bibitem[\protect\citeauthoryear{Ren et~al.}{2024}]{topnav}
\begin{botherref}
\oauthor{\bsnm{Ren}, \binits{J.}},
\oauthor{\bsnm{Liu}, \binits{Y.}},
\oauthor{\bsnm{Dai}, \binits{Y.}},
\oauthor{\bsnm{Wang}, \binits{G.}}:
Top-nav: Legged navigation integrating terrain, obstacle and proprioception estimation.
arXiv preprint arXiv:2404.15256
(2024)
\end{botherref}
\endbibitem

\bibitem[\protect\citeauthoryear{Fu et~al.}{2022}]{coupling}
\begin{bchapter}
\bauthor{\bsnm{Fu}, \binits{Z.}},
\bauthor{\bsnm{Kumar}, \binits{A.}},
\bauthor{\bsnm{Agarwal}, \binits{A.}},
\bauthor{\bsnm{Qi}, \binits{H.}},
\bauthor{\bsnm{Malik}, \binits{J.}},
\bauthor{\bsnm{Pathak}, \binits{D.}}:
\bctitle{Coupling vision and proprioception for navigation of legged robots}.
In: \bbtitle{Proceedings of the IEEE/CVF Conference on Computer Vision and Pattern Recognition},
pp. \bfpage{17273}--\blpage{17283}
(\byear{2022})
\end{bchapter}
\endbibitem

\bibitem[\protect\citeauthoryear{Karnan et~al.}{2023}]{karnan2023sterling}
\begin{bchapter}
\bauthor{\bsnm{Karnan}, \binits{H.}},
\bauthor{\bsnm{Yang}, \binits{E.}},
\bauthor{\bsnm{Farkash}, \binits{D.}},
\bauthor{\bsnm{Warnell}, \binits{G.}},
\bauthor{\bsnm{Biswas}, \binits{J.}},
\bauthor{\bsnm{Stone}, \binits{P.}}:
\bctitle{Sterling: Self-supervised terrain representation learning from unconstrained robot experience}.
In: \bbtitle{7th Annual Conference on Robot Learning}
(\byear{2023})
\end{bchapter}
\endbibitem

\bibitem[\protect\citeauthoryear{Truong et~al.}{2021}]{2021learningembeddings}
\begin{bchapter}
\bauthor{\bsnm{Truong}, \binits{J.}},
\bauthor{\bsnm{Yarats}, \binits{D.}},
\bauthor{\bsnm{Li}, \binits{T.}},
\bauthor{\bsnm{Meier}, \binits{F.}},
\bauthor{\bsnm{Chernova}, \binits{S.}},
\bauthor{\bsnm{Batra}, \binits{D.}},
\bauthor{\bsnm{Rai}, \binits{A.}}:
\bctitle{Learning navigation skills for legged robots with learned robot embeddings}.
In: \bbtitle{2021 IEEE/RSJ International Conference on Intelligent Robots and Systems (IROS)},
pp. \bfpage{484}--\blpage{491}
(\byear{2021}).
\bcomment{IEEE}
\end{bchapter}
\endbibitem

\bibitem[\protect\citeauthoryear{Cheng et~al.}{2024}]{extreme}
\begin{bchapter}
\bauthor{\bsnm{Cheng}, \binits{X.}},
\bauthor{\bsnm{Shi}, \binits{K.}},
\bauthor{\bsnm{Agarwal}, \binits{A.}},
\bauthor{\bsnm{Pathak}, \binits{D.}}:
\bctitle{Extreme parkour with legged robots}.
In: \bbtitle{2024 IEEE International Conference on Robotics and Automation (ICRA)},
pp. \bfpage{11443}--\blpage{11450}
(\byear{2024}).
\bcomment{IEEE}
\end{bchapter}
\endbibitem

\bibitem[\protect\citeauthoryear{Zhuang et~al.}{2023}]{robotparkour}
\begin{bchapter}
\bauthor{\bsnm{Zhuang}, \binits{Z.}},
\bauthor{\bsnm{Fu}, \binits{Z.}},
\bauthor{\bsnm{Wang}, \binits{J.}},
\bauthor{\bsnm{Atkeson}, \binits{C.}},
\bauthor{\bsnm{Schwertfeger}, \binits{S.}},
\bauthor{\bsnm{Finn}, \binits{C.}},
\bauthor{\bsnm{Zhao}, \binits{H.}}:
\bctitle{Robot parkour learning}.
In: \bbtitle{Conference on Robot Learning ({CoRL})}
(\byear{2023})
\end{bchapter}
\endbibitem

\bibitem[\protect\citeauthoryear{Barasuol et~al.}{2024}]{2024High-Obstacle}
\begin{bchapter}
\bauthor{\bsnm{Barasuol}, \binits{V.}},
\bauthor{\bsnm{Emre}, \binits{S.}},
\bauthor{\bsnm{Medeiros}, \binits{V.S.}},
\bauthor{\bsnm{Bratta}, \binits{A.}},
\bauthor{\bsnm{Semini}, \binits{C.}}:
\bctitle{Introducing the carpal-claw: a mechanism to enhance high-obstacle negotiation for quadruped robots}.
In: \bbtitle{2024 IEEE International Conference on Robotics and Automation (ICRA)},
pp. \bfpage{3457}--\blpage{3463}
(\byear{2024}).
\bcomment{IEEE}
\end{bchapter}
\endbibitem

\bibitem[\protect\citeauthoryear{He et~al.}{2024}]{agilebufsafe}
\begin{bchapter}
\bauthor{\bsnm{He}, \binits{T.}},
\bauthor{\bsnm{Zhang}, \binits{C.}},
\bauthor{\bsnm{Xiao}, \binits{W.}},
\bauthor{\bsnm{He}, \binits{G.}},
\bauthor{\bsnm{Liu}, \binits{C.}},
\bauthor{\bsnm{Shi}, \binits{G.}}:
\bctitle{Agile but safe: Learning collision-free high-speed legged locomotion}.
In: \bbtitle{Robotics: Science and Systems (RSS)}
(\byear{2024})
\end{bchapter}
\endbibitem

\bibitem[\protect\citeauthoryear{Lee et~al.}{2024}]{wheeled-legged}
\begin{barticle}
\bauthor{\bsnm{Lee}, \binits{J.}},
\bauthor{\bsnm{Bjelonic}, \binits{M.}},
\bauthor{\bsnm{Reske}, \binits{A.}},
\bauthor{\bsnm{Wellhausen}, \binits{L.}},
\bauthor{\bsnm{Miki}, \binits{T.}},
\bauthor{\bsnm{Hutter}, \binits{M.}}:
\batitle{Learning robust autonomous navigation and locomotion for wheeled-legged robots}.
\bjtitle{Science Robotics}
\bvolume{9}(\bissue{89}),
\bfpage{9641}
(\byear{2024})
\end{barticle}
\endbibitem

\bibitem[\protect\citeauthoryear{Hoeller et~al.}{2024}]{Anymalparkour}
\begin{barticle}
\bauthor{\bsnm{Hoeller}, \binits{D.}},
\bauthor{\bsnm{Rudin}, \binits{N.}},
\bauthor{\bsnm{Sako}, \binits{D.}},
\bauthor{\bsnm{Hutter}, \binits{M.}}:
\batitle{Anymal parkour: Learning agile navigation for quadrupedal robots}.
\bjtitle{Science Robotics}
\bvolume{9}(\bissue{88}),
\bfpage{7566}
(\byear{2024})
\end{barticle}
\endbibitem

\bibitem[\protect\citeauthoryear{Grandia et~al.}{2023}]{2023MPC1}
\begin{barticle}
\bauthor{\bsnm{Grandia}, \binits{R.}},
\bauthor{\bsnm{Jenelten}, \binits{F.}},
\bauthor{\bsnm{Yang}, \binits{S.}},
\bauthor{\bsnm{Farshidian}, \binits{F.}},
\bauthor{\bsnm{Hutter}, \binits{M.}}:
\batitle{Perceptive locomotion through nonlinear model-predictive control}.
\bjtitle{IEEE Transactions on Robotics}
\bvolume{39}(\bissue{5}),
\bfpage{3402}--\blpage{3421}
(\byear{2023})
\end{barticle}
\endbibitem

\bibitem[\protect\citeauthoryear{Margolis and Agrawal}{2023}]{2023walkthese}
\begin{bchapter}
\bauthor{\bsnm{Margolis}, \binits{G.B.}},
\bauthor{\bsnm{Agrawal}, \binits{P.}}:
\bctitle{Walk these ways: Tuning robot control for generalization with multiplicity of behavior}.
In: \bbtitle{Conference on Robot Learning},
pp. \bfpage{22}--\blpage{31}
(\byear{2023}).
\bcomment{PMLR}
\end{bchapter}
\endbibitem

\bibitem[\protect\citeauthoryear{Margolis et~al.}{2024}]{2024rapid}
\begin{barticle}
\bauthor{\bsnm{Margolis}, \binits{G.B.}},
\bauthor{\bsnm{Yang}, \binits{G.}},
\bauthor{\bsnm{Paigwar}, \binits{K.}},
\bauthor{\bsnm{Chen}, \binits{T.}},
\bauthor{\bsnm{Agrawal}, \binits{P.}}:
\batitle{Rapid locomotion via reinforcement learning}.
\bjtitle{The International Journal of Robotics Research}
\bvolume{43}(\bissue{4}),
\bfpage{572}--\blpage{587}
(\byear{2024})
\end{barticle}
\endbibitem

\bibitem[\protect\citeauthoryear{Shi et~al.}{2024}]{shi2024robust}
\begin{bchapter}
\bauthor{\bsnm{Shi}, \binits{J.}},
\bauthor{\bsnm{Bai}, \binits{C.}},
\bauthor{\bsnm{He}, \binits{H.}},
\bauthor{\bsnm{Han}, \binits{L.}},
\bauthor{\bsnm{Wang}, \binits{D.}},
\bauthor{\bsnm{Zhao}, \binits{B.}},
\bauthor{\bsnm{Zhao}, \binits{M.}},
\bauthor{\bsnm{Li}, \binits{X.}},
\bauthor{\bsnm{Li}, \binits{X.}}:
\bctitle{Robust quadrupedal locomotion via risk-averse policy learning}.
In: \bbtitle{2024 IEEE International Conference on Robotics and Automation (ICRA)},
pp. \bfpage{11459}--\blpage{11466}
(\byear{2024}).
\bcomment{IEEE}
\end{bchapter}
\endbibitem

\bibitem[\protect\citeauthoryear{Long et~al.}{2024}]{2024HIM}
\begin{bchapter}
\bauthor{\bsnm{Long}, \binits{J.}},
\bauthor{\bsnm{Wang}, \binits{Z.}},
\bauthor{\bsnm{Li}, \binits{Q.}},
\bauthor{\bsnm{Cao}, \binits{L.}},
\bauthor{\bsnm{Gao}, \binits{J.}},
\bauthor{\bsnm{Pang}, \binits{J.}}:
\bctitle{Hybrid internal model: Learning agile legged locomotion with simulated robot response}.
In: \bbtitle{The Twelfth International Conference on Learning Representations}
(\byear{2024})
\end{bchapter}
\endbibitem

\bibitem[\protect\citeauthoryear{Cheng et~al.}{2024}]{2024limitedperception}
\begin{botherref}
\oauthor{\bsnm{Cheng}, \binits{Y.}},
\oauthor{\bsnm{Liu}, \binits{H.}},
\oauthor{\bsnm{Pan}, \binits{G.}},
\oauthor{\bsnm{Ye}, \binits{L.}},
\oauthor{\bsnm{Liu}, \binits{H.}},
\oauthor{\bsnm{Liang}, \binits{B.}}:
Quadruped robot traversing 3d complex environments with limited perception.
arXiv preprint arXiv:2404.18225
(2024)
\end{botherref}
\endbibitem

\bibitem[\protect\citeauthoryear{Shi et~al.}{2024}]{shi2024rethinking}
\begin{botherref}
\oauthor{\bsnm{Shi}, \binits{F.}},
\oauthor{\bsnm{Zhang}, \binits{C.}},
\oauthor{\bsnm{Miki}, \binits{T.}},
\oauthor{\bsnm{Lee}, \binits{J.}},
\oauthor{\bsnm{Hutter}, \binits{M.}},
\oauthor{\bsnm{Coros}, \binits{S.}}:
Rethinking robustness assessment: Adversarial attacks on learning-based quadrupedal locomotion controllers.
arXiv preprint arXiv:2405.12424
(2024)
\end{botherref}
\endbibitem

\bibitem[\protect\citeauthoryear{Yang et~al.}{2022}]{2022lquaTransformers}
\begin{bchapter}
\bauthor{\bsnm{Yang}, \binits{R.}},
\bauthor{\bsnm{Zhang}, \binits{M.}},
\bauthor{\bsnm{Hansen}, \binits{N.}},
\bauthor{\bsnm{Xu}, \binits{H.}},
\bauthor{\bsnm{Wang}, \binits{X.}}:
\bctitle{Learning vision-guided quadrupedal locomotion end-to-end with cross-modal transformers}.
In: \bbtitle{International Conference on Learning Representations}
(\byear{2022}).
\burl{https://openreview.net/forum?id=nhnJ3oo6AB}
\end{bchapter}
\endbibitem

\bibitem[\protect\citeauthoryear{Miki et~al.}{2024}]{miki2024confined}
\begin{botherref}
\oauthor{\bsnm{Miki}, \binits{T.}},
\oauthor{\bsnm{Lee}, \binits{J.}},
\oauthor{\bsnm{Wellhausen}, \binits{L.}},
\oauthor{\bsnm{Hutter}, \binits{M.}}:
Learning to walk in confined spaces using 3d representation.
arXiv preprint arXiv:2403.00187
(2024)
\end{botherref}
\endbibitem

\bibitem[\protect\citeauthoryear{Agarwal et~al.}{2023}]{2023egocentric}
\begin{bchapter}
\bauthor{\bsnm{Agarwal}, \binits{A.}},
\bauthor{\bsnm{Kumar}, \binits{A.}},
\bauthor{\bsnm{Malik}, \binits{J.}},
\bauthor{\bsnm{Pathak}, \binits{D.}}:
\bctitle{Legged locomotion in challenging terrains using egocentric vision}.
In: \bbtitle{Conference on Robot Learning},
pp. \bfpage{403}--\blpage{415}
(\byear{2023}).
\bcomment{PMLR}
\end{bchapter}
\endbibitem

\bibitem[\protect\citeauthoryear{Seo et~al.}{2023}]{walkbysteering}
\begin{bchapter}
\bauthor{\bsnm{Seo}, \binits{M.}},
\bauthor{\bsnm{Gupta}, \binits{R.}},
\bauthor{\bsnm{Zhu}, \binits{Y.}},
\bauthor{\bsnm{Skoutnev}, \binits{A.}},
\bauthor{\bsnm{Sentis}, \binits{L.}},
\bauthor{\bsnm{Zhu}, \binits{Y.}}:
\bctitle{Learning to walk by steering: Perceptive quadrupedal locomotion in dynamic environments}.
In: \bbtitle{2023 IEEE International Conference on Robotics and Automation (ICRA)},
pp. \bfpage{5099}--\blpage{5105}
(\byear{2023}).
\bcomment{IEEE}
\end{bchapter}
\endbibitem

\bibitem[\protect\citeauthoryear{Jenelten et~al.}{2023}]{DTC}
\begin{botherref}
\oauthor{\bsnm{Jenelten}, \binits{F.}},
\oauthor{\bsnm{He}, \binits{J.}},
\oauthor{\bsnm{Farshidian}, \binits{F.}},
\oauthor{\bsnm{Hutter}, \binits{M.}}:
Dtc: Deep tracking control--a unifying approach to model-based planning and reinforcement-learning for versatile and robust locomotion.
arXiv preprint arXiv:2309.15462
(2023)
\end{botherref}
\endbibitem

\bibitem[\protect\citeauthoryear{Kang et~al.}{2023}]{kang2023rl+}
\begin{botherref}
\oauthor{\bsnm{Kang}, \binits{D.}},
\oauthor{\bsnm{Cheng}, \binits{J.}},
\oauthor{\bsnm{Zamora}, \binits{M.}},
\oauthor{\bsnm{Zargarbashi}, \binits{F.}},
\oauthor{\bsnm{Coros}, \binits{S.}}:
Rl+ model-based control: Using on-demand optimal control to learn versatile legged locomotion.
IEEE Robotics and Automation Letters
(2023)
\end{botherref}
\endbibitem

\bibitem[\protect\citeauthoryear{Rudin et~al.}{2022}]{advancedskills}
\begin{bchapter}
\bauthor{\bsnm{Rudin}, \binits{N.}},
\bauthor{\bsnm{Hoeller}, \binits{D.}},
\bauthor{\bsnm{Bjelonic}, \binits{M.}},
\bauthor{\bsnm{Hutter}, \binits{M.}}:
\bctitle{Advanced skills by learning locomotion and local navigation end-to-end}.
In: \bbtitle{2022 IEEE/RSJ International Conference on Intelligent Robots and Systems (IROS)},
pp. \bfpage{2497}--\blpage{2503}
(\byear{2022}).
\bcomment{IEEE}
\end{bchapter}
\endbibitem

\bibitem[\protect\citeauthoryear{Zhang et~al.}{2024}]{resilient}
\begin{bchapter}
\bauthor{\bsnm{Zhang}, \binits{C.}},
\bauthor{\bsnm{Jin}, \binits{J.}},
\bauthor{\bsnm{Frey}, \binits{J.}},
\bauthor{\bsnm{Rudin}, \binits{N.}},
\bauthor{\bsnm{Mattamala}, \binits{M.}},
\bauthor{\bsnm{Cadena}, \binits{C.}},
\bauthor{\bsnm{Hutter}, \binits{M.}}:
\bctitle{Resilient legged local navigation: Learning to traverse with compromised perception end-to-end}.
In: \bbtitle{2024 IEEE International Conference on Robotics and Automation (ICRA)},
pp. \bfpage{34}--\blpage{41}
(\byear{2024}).
\bcomment{IEEE}
\end{bchapter}
\endbibitem

\bibitem[\protect\citeauthoryear{Zhang et~al.}{2023}]{2023risky}
\begin{botherref}
\oauthor{\bsnm{Zhang}, \binits{C.}},
\oauthor{\bsnm{Rudin}, \binits{N.}},
\oauthor{\bsnm{Hoeller}, \binits{D.}},
\oauthor{\bsnm{Hutter}, \binits{M.}}:
Learning agile locomotion on risky terrains.
arXiv preprint arXiv:2311.10484
(2023)
\end{botherref}
\endbibitem

\bibitem[\protect\citeauthoryear{Cheng et~al.}{2024}]{2024learningdiverse}
\begin{bchapter}
\bauthor{\bsnm{Cheng}, \binits{J.}},
\bauthor{\bsnm{Vlastelica}, \binits{M.}},
\bauthor{\bsnm{Kolev}, \binits{P.}},
\bauthor{\bsnm{Li}, \binits{C.}},
\bauthor{\bsnm{Martius}, \binits{G.}}:
\bctitle{Learning diverse skills for local navigation under multi-constraint optimality}.
In: \bbtitle{2024 IEEE International Conference on Robotics and Automation (ICRA)},
pp. \bfpage{5083}--\blpage{5089}
(\byear{2024}).
\bcomment{IEEE}
\end{bchapter}
\endbibitem

\bibitem[\protect\citeauthoryear{Faust et~al.}{2018}]{faust2018prmrl}
\begin{bchapter}
\bauthor{\bsnm{Faust}, \binits{A.}},
\bauthor{\bsnm{Oslund}, \binits{K.}},
\bauthor{\bsnm{Ramirez}, \binits{O.}},
\bauthor{\bsnm{Francis}, \binits{A.}},
\bauthor{\bsnm{Tapia}, \binits{L.}},
\bauthor{\bsnm{Fiser}, \binits{M.}},
\bauthor{\bsnm{Davidson}, \binits{J.}}:
\bctitle{Prm-rl: Long-range robotic navigation tasks by combining reinforcement learning and sampling-based planning}.
In: \bbtitle{2018 IEEE International Conference on Robotics and Automation (ICRA)},
pp. \bfpage{5113}--\blpage{5120}
(\byear{2018}).
\bcomment{IEEE}
\end{bchapter}
\endbibitem

\bibitem[\protect\citeauthoryear{Xiao et~al.}{2022}]{xiao2022motion}
\begin{barticle}
\bauthor{\bsnm{Xiao}, \binits{X.}},
\bauthor{\bsnm{Liu}, \binits{B.}},
\bauthor{\bsnm{Warnell}, \binits{G.}},
\bauthor{\bsnm{Stone}, \binits{P.}}:
\batitle{Motion planning and control for mobile robot navigation using machine learning: a survey}.
\bjtitle{Autonomous Robots}
\bvolume{46}(\bissue{5}),
\bfpage{569}--\blpage{597}
(\byear{2022})
\end{barticle}
\endbibitem

\bibitem[\protect\citeauthoryear{Wellhausen and Hutter}{2021}]{wellhausen2021rough}
\begin{bchapter}
\bauthor{\bsnm{Wellhausen}, \binits{L.}},
\bauthor{\bsnm{Hutter}, \binits{M.}}:
\bctitle{Rough terrain navigation for legged robots using reachability planning and template learning}.
In: \bbtitle{2021 IEEE/RSJ International Conference on Intelligent Robots and Systems (IROS)},
pp. \bfpage{6914}--\blpage{6921}
(\byear{2021}).
\bcomment{IEEE}
\end{bchapter}
\endbibitem

\bibitem[\protect\citeauthoryear{Pfeiffer et~al.}{2017}]{pfeiffer2017perception}
\begin{bchapter}
\bauthor{\bsnm{Pfeiffer}, \binits{M.}},
\bauthor{\bsnm{Schaeuble}, \binits{M.}},
\bauthor{\bsnm{Nieto}, \binits{J.}},
\bauthor{\bsnm{Siegwart}, \binits{R.}},
\bauthor{\bsnm{Cadena}, \binits{C.}}:
\bctitle{From perception to decision: A data-driven approach to end-to-end motion planning for autonomous ground robots}.
In: \bbtitle{2017 Ieee International Conference on Robotics and Automation (icra)},
pp. \bfpage{1527}--\blpage{1533}
(\byear{2017}).
\bcomment{IEEE}
\end{bchapter}
\endbibitem

\bibitem[\protect\citeauthoryear{Zhu et~al.}{2017}]{zhu2017target}
\begin{bchapter}
\bauthor{\bsnm{Zhu}, \binits{Y.}},
\bauthor{\bsnm{Mottaghi}, \binits{R.}},
\bauthor{\bsnm{Kolve}, \binits{E.}},
\bauthor{\bsnm{Lim}, \binits{J.J.}},
\bauthor{\bsnm{Gupta}, \binits{A.}},
\bauthor{\bsnm{Fei-Fei}, \binits{L.}},
\bauthor{\bsnm{Farhadi}, \binits{A.}}:
\bctitle{Target-driven visual navigation in indoor scenes using deep reinforcement learning}.
In: \bbtitle{2017 IEEE International Conference on Robotics and Automation (ICRA)},
pp. \bfpage{3357}--\blpage{3364}
(\byear{2017}).
\bcomment{IEEE}
\end{bchapter}
\endbibitem

\bibitem[\protect\citeauthoryear{Shah et~al.}{2023}]{shah2023vint}
\begin{bchapter}
\bauthor{\bsnm{Shah}, \binits{D.}},
\bauthor{\bsnm{Sridhar}, \binits{A.}},
\bauthor{\bsnm{Dashora}, \binits{N.}},
\bauthor{\bsnm{Stachowicz}, \binits{K.}},
\bauthor{\bsnm{Black}, \binits{K.}},
\bauthor{\bsnm{Hirose}, \binits{N.}},
\bauthor{\bsnm{Levine}, \binits{S.}}:
\bctitle{Vi{NT}: A foundation model for visual navigation}.
In: \bbtitle{7th Annual Conference on Robot Learning}
(\byear{2023}).
\burl{https://arxiv.org/abs/2306.14846}
\end{bchapter}
\endbibitem

\bibitem[\protect\citeauthoryear{Gao et~al.}{2024}]{IntentionNet}
\begin{botherref}
\oauthor{\bsnm{Gao}, \binits{W.}},
\oauthor{\bsnm{Ai}, \binits{B.}},
\oauthor{\bsnm{Loo}, \binits{J.}},
\oauthor{\bsnm{Hsu}, \binits{D.}}, et al.:
Intentionnet: Map-lite visual navigation at the kilometre scale.
arXiv preprint arXiv:2407.03122
(2024)
\end{botherref}
\endbibitem

\bibitem[\protect\citeauthoryear{Wang et~al.}{2024}]{2024quadrupedgpt}
\begin{botherref}
\oauthor{\bsnm{Wang}, \binits{Y.}},
\oauthor{\bsnm{Mei}, \binits{Y.}},
\oauthor{\bsnm{Zheng}, \binits{S.}},
\oauthor{\bsnm{Jin}, \binits{Q.}}:
Quadrupedgpt: Towards a versatile quadruped agent in open-ended worlds.
arXiv preprint arXiv:2406.16578
(2024)
\end{botherref}
\endbibitem

\bibitem[\protect\citeauthoryear{Chen et~al.}{2024}]{commonsense}
\begin{botherref}
\oauthor{\bsnm{Chen}, \binits{A.S.}},
\oauthor{\bsnm{Lessing}, \binits{A.M.}},
\oauthor{\bsnm{Tang}, \binits{A.}},
\oauthor{\bsnm{Chada}, \binits{G.}},
\oauthor{\bsnm{Smith}, \binits{L.}},
\oauthor{\bsnm{Levine}, \binits{S.}},
\oauthor{\bsnm{Finn}, \binits{C.}}:
Commonsense reasoning for legged robot adaptation with vision-language models.
arXiv preprint arXiv:2407.02666
(2024)
\end{botherref}
\endbibitem

\bibitem[\protect\citeauthoryear{Schulman et~al.}{2017}]{2017ppo}
\begin{botherref}
\oauthor{\bsnm{Schulman}, \binits{J.}},
\oauthor{\bsnm{Wolski}, \binits{F.}},
\oauthor{\bsnm{Dhariwal}, \binits{P.}},
\oauthor{\bsnm{Radford}, \binits{A.}},
\oauthor{\bsnm{Klimov}, \binits{O.}}:
Proximal policy optimization algorithms.
arXiv preprint arXiv:1707.06347
(2017)
\end{botherref}
\endbibitem

\bibitem[\protect\citeauthoryear{Driess et~al.}{2023}]{palm_e}
\begin{bchapter}
\bauthor{\bsnm{Driess}, \binits{D.}},
\bauthor{\bsnm{Xia}, \binits{F.}},
\bauthor{\bsnm{Sajjadi}, \binits{M.S.M.}},
\bauthor{\bsnm{Lynch}, \binits{C.}},
\bauthor{\bsnm{Chowdhery}, \binits{A.}},
\bauthor{\bsnm{Ichter}, \binits{B.}},
\bauthor{\bsnm{Wahid}, \binits{A.}},
\bauthor{\bsnm{Tompson}, \binits{J.}},
\bauthor{\bsnm{Vuong}, \binits{Q.}},
\bauthor{\bsnm{Yu}, \binits{T.}},
\bauthor{\bsnm{Huang}, \binits{W.}},
\bauthor{\bsnm{Chebotar}, \binits{Y.}},
\bauthor{\bsnm{Sermanet}, \binits{P.}},
\bauthor{\bsnm{Duckworth}, \binits{D.}},
\bauthor{\bsnm{Levine}, \binits{S.}},
\bauthor{\bsnm{Vanhoucke}, \binits{V.}},
\bauthor{\bsnm{Hausman}, \binits{K.}},
\bauthor{\bsnm{Toussaint}, \binits{M.}},
\bauthor{\bsnm{Greff}, \binits{K.}},
\bauthor{\bsnm{Zeng}, \binits{A.}},
\bauthor{\bsnm{Mordatch}, \binits{I.}},
\bauthor{\bsnm{Florence}, \binits{P.}}:
\bctitle{Palm-e: An embodied multimodal language model}.
In: \bbtitle{arXiv Preprint arXiv:2303.03378}
(\byear{2023})
\end{bchapter}
\endbibitem

\bibitem[\protect\citeauthoryear{Li}{2022}]{li2022positive}
\begin{barticle}
\bauthor{\bsnm{Li}, \binits{X.}}:
\batitle{Positive-incentive noise}.
\bjtitle{IEEE Transactions on Neural Networks and Learning Systems}
\bvolume{35}(\bissue{6}),
\bfpage{8708}--\blpage{8714}
(\byear{2022})
\end{barticle}
\endbibitem

\bibitem[\protect\citeauthoryear{Zhang et~al.}{2025}]{zhang2025variational}
\begin{botherref}
\oauthor{\bsnm{Zhang}, \binits{H.}},
\oauthor{\bsnm{Huang}, \binits{S.}},
\oauthor{\bsnm{Guo}, \binits{Y.}},
\oauthor{\bsnm{Li}, \binits{X.}}:
Variational positive-incentive noise: How noise benefits models.
IEEE Transactions on Pattern Analysis and Machine Intelligence
(2025)
\end{botherref}
\endbibitem

\bibitem[\protect\citeauthoryear{Huang et~al.}{2025}]{huang2025enhance}
\begin{bchapter}
\bauthor{\bsnm{Huang}, \binits{S.}},
\bauthor{\bsnm{Zhang}, \binits{H.}},
\bauthor{\bsnm{Li}, \binits{X.}}:
\bctitle{Enhance vision-language alignment with noise}.
In: \bbtitle{Proceedings of the AAAI Conference on Artificial Intelligence},
vol. \bseriesno{39},
pp. \bfpage{17449}--\blpage{17457}
(\byear{2025})
\end{bchapter}
\endbibitem

\bibitem[\protect\citeauthoryear{Zhang et~al.}{2024}]{zhang2024data}
\begin{botherref}
\oauthor{\bsnm{Zhang}, \binits{H.}},
\oauthor{\bsnm{Xu}, \binits{Y.}},
\oauthor{\bsnm{Huang}, \binits{S.}},
\oauthor{\bsnm{Li}, \binits{X.}}:
Data augmentation of contrastive learning is estimating positive-incentive noise.
arXiv preprint arXiv:2408.09929
(2024)
\end{botherref}
\endbibitem

\bibitem[\protect\citeauthoryear{Huang et~al.}{2025}]{huang2025learn}
\begin{botherref}
\oauthor{\bsnm{Huang}, \binits{S.}},
\oauthor{\bsnm{Xu}, \binits{Y.}},
\oauthor{\bsnm{Zhang}, \binits{H.}},
\oauthor{\bsnm{Li}, \binits{X.}}:
Learn beneficial noise as graph augmentation.
arXiv preprint arXiv:2505.19024
(2025)
\end{botherref}
\endbibitem

\bibitem[\protect\citeauthoryear{Achiam et~al.}{2023}]{achiam2023gpt}
\begin{botherref}
\oauthor{\bsnm{Achiam}, \binits{J.}},
\oauthor{\bsnm{Adler}, \binits{S.}},
\oauthor{\bsnm{Agarwal}, \binits{S.}},
\oauthor{\bsnm{Ahmad}, \binits{L.}},
\oauthor{\bsnm{Akkaya}, \binits{I.}},
\oauthor{\bsnm{Aleman}, \binits{F.L.}},
\oauthor{\bsnm{Almeida}, \binits{D.}},
\oauthor{\bsnm{Altenschmidt}, \binits{J.}},
\oauthor{\bsnm{Altman}, \binits{S.}},
\oauthor{\bsnm{Anadkat}, \binits{S.}}, et al.:
Gpt-4 technical report.
arXiv preprint arXiv:2303.08774
(2023)
\end{botherref}
\endbibitem

\end{thebibliography}

\end{document}